\documentclass[twocolumn,letterpaper]{IEEEAerospaceCLS}  

\usepackage[]{graphicx}    
\usepackage{gensymb}
\usepackage{float}
\usepackage{xcolor}
\usepackage[noadjust]{cite} 
\usepackage{enumitem}
\usepackage{array}\usepackage{url}
\usepackage{graphicx, caption, subcaption}

\newcommand{\ignore}[1]{}  

\begin{document}
\title{Lunar Rover Localization Using Craters as Landmarks}

\author{%
Larry Matthies, Shreyansh Daftry, Scott Tepsuporn, Yang Cheng, Deegan Atha, R. Michael Swan \\
Sanjna Ravichandar and Masahiro Ono\\
Jet propulsion Laboratory, California Institute of Technology\\
4800 Oak Grove Drive\\
Pasadena, CA 90245\\
lhm@jpl.nasa.gov
\thanks{\footnotesize 978-1-6654-3760-8/22/$\$31.00$ \copyright2022 IEEE}              
}

\maketitle

\thispagestyle{plain}
\pagestyle{plain}

\maketitle

\thispagestyle{plain}
\pagestyle{plain}

\begin{abstract}
Onboard localization capabilities for planetary rovers to date have used relative navigation, by integrating combinations of wheel odometry, visual odometry, and inertial measurements during each drive to track position relative to the start of each drive. At the end of each drive, a “ground-in-the-loop” (GITL) interaction is used to get a position update from human operators in a more global reference frame, by matching images or local maps from onboard the rover to orbital reconnaissance images or maps of a large region around the rover’s current position. Autonomous rover drives are limited in distance so that accumulated relative navigation error does not risk the possibility of the rover driving into hazards known from orbital images. In practice, this limits drives to a few hundred meters between GITL cycles. Several rover mission concepts have recently been studied that require much longer drives between GITL cycles, particularly for the Moon. This includes lunar rover mission concepts that involve (1) driving mostly in sunlight at low latitudes, (2) driving in permanently shadowed regions near the south pole, and (3) a mixture of day and night driving in mid-latitudes. These concepts include total traverse distance requirements of up to 1,800 km in 4 Earth years, with individual drives of several kilometers between stops for downlink. These concepts require greater autonomy to minimize GITL cycles to enable such large range; onboard global localization is a key element of such autonomy. Multiple techniques have been studied in the past for onboard rover global localization, but a satisfactory solution has not yet emerged. For the Moon, the ubiquitous craters offer a new possibility, which involves mapping craters from orbit, then recognizing crater landmarks with cameras and/or a lidar onboard the rover. This approach is applicable everywhere on the Moon, does not require high resolution stereo imaging from orbit as some other approaches do, and has potential to enable position knowledge with order of 5 to 10 m accuracy at all times. This paper describes our technical approach to crater-based lunar rover localization and presents initial results on crater detection using 3-D point cloud data from onboard lidar or stereo cameras, as well as using shading cues in monocular onboard imagery.

\end{abstract}

\tableofcontents

\section{Introduction}
\label{}
Lunar surface exploration is seeing a renaissance, driven by work toward returning humans to the Moon \cite{smith2020artemis} and by new interest in rover-based lunar science investigations \cite{cohen2020lunar}. Some of these mission concepts require extremely long traverses; in particular, the Intrepid rover mission concept specifies a traverse of ~ 1,800 km over four Earth years, driving mostly in sunlight at relative low latitude on the lunar near side \cite{robinson2019intrepid}. Other mission concepts involve extensive driving in areas near the south pole that are frequently or permanently in shadow, such as the VIPER mission now in development \cite{colaprete2019overview} and the Lunar Polar Volatiles Explorer mission concept \cite{cohen2020lunar}.

All robotic rover missions require knowledge of the rover’s incremental position change during each driving segment (relative localization) and absolute position in a regional map reference frame (absolute localization) for navigation planning. Relative localization has been done effectively onboard by Mars rovers, using a combination of wheel odometry, inertial measurement units (IMUs), and visual odometry for deadreckoning, plus sun sensing for absolute heading updates \cite{ali2005attitude}. All rover missions that have flown to date have done absolute localization with assistance from Earth by stopping to downlink images, which human operators interactively register to orbital imagery to estimate rover position. For long-range rover missions that must maintain fast progress over a long duration, such “ground-in-the-loop” (GITL) localization methods are impractical, because they either take too long, are too demanding of the human operations team, or both. New methods are needed that provide onboard, automated absolute localization and eliminate the need for GITL cycles.

\begin{figure}[!t]
    \centering
    \includegraphics[width=\linewidth]{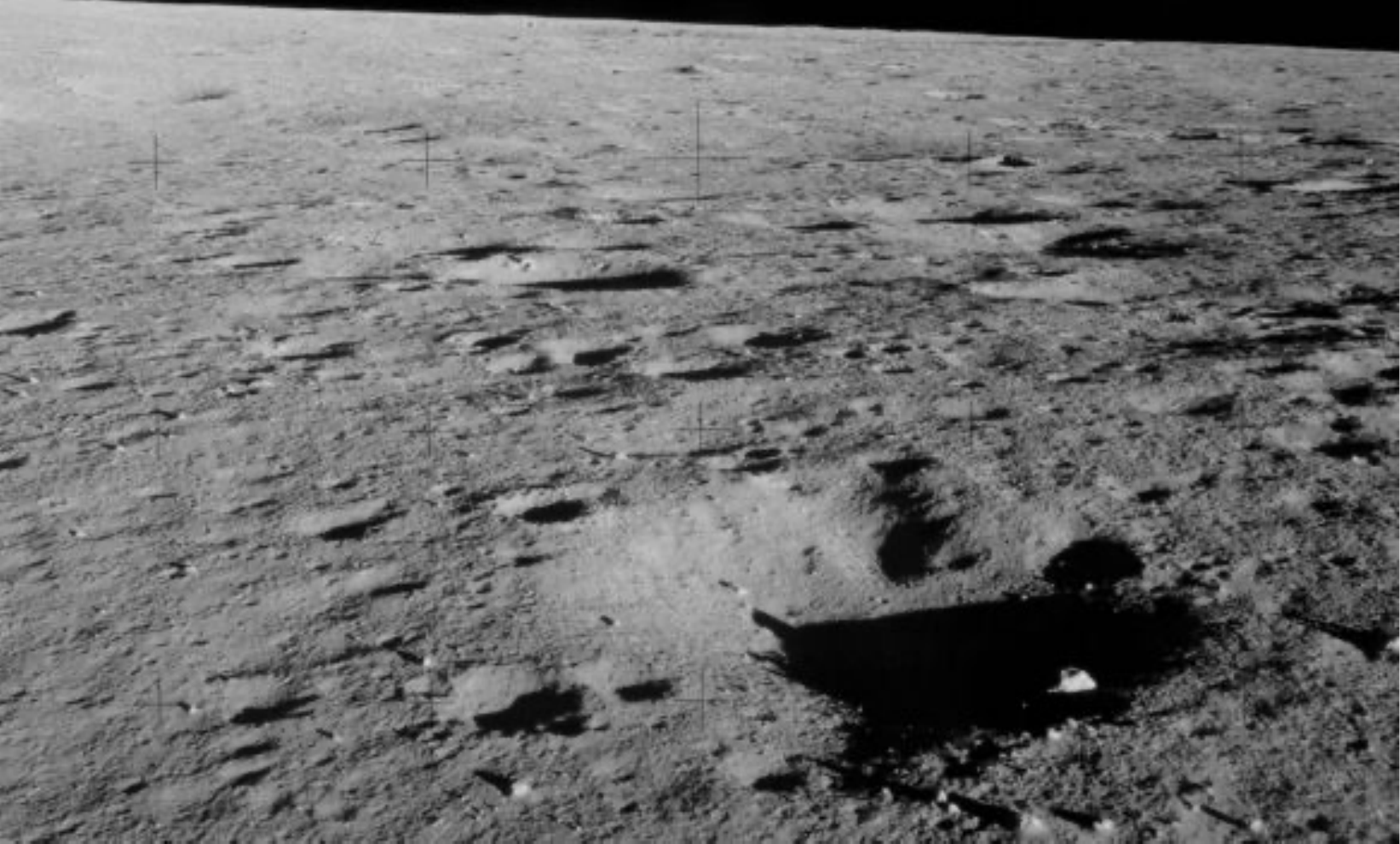}
    \caption{\textbf{Lunar surface image taken by the Apollo mission illustrating the frequency of lunar craters.}}
    \label{fig:teaser}
\end{figure}

\begin{figure*}[!t]
    \centering
    \includegraphics[width=\linewidth]{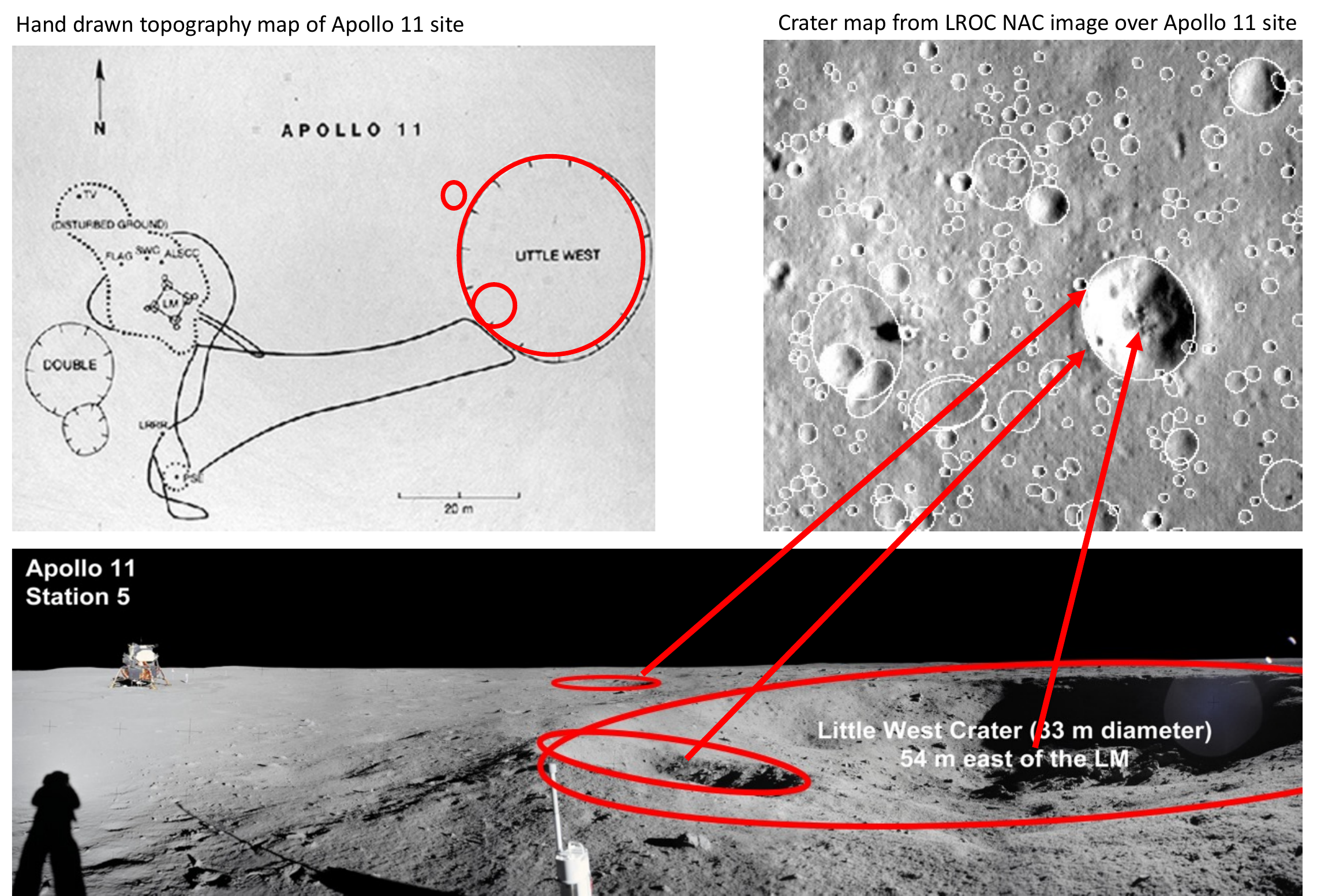}
    \caption{\textbf{Illustration of LunarNav's overall system concept to recognize craters from  surface images and match to the crater map will help to localize rover globally}}
    \label{fig:system_concept}
\end{figure*}

A number of approaches have been studied for onboard absolute localization, including creating local elevation maps onboard to register with elevation maps created from orbit, matching visual keypoints extracted from onboard images to keypoints extracted from orbital images, using horizon features as landmarks, and other methods. These all have drawbacks and none have reached sufficient maturity to integrate into a mission. For lunar rovers, a new approach is to use craters as landmarks. Craters are ubiquitous and plentiful on the Moon (see Figure \ref{fig:teaser}) \cite{hiesinger2012old}. A very high percentage of the Moon has been mapped at resolutions up to 50 cm/pixel by the Narrow Angle instrument on the Lunar Reconnaissance Orbiter Camera (LROC-NA) \cite{robinson2010lunar}. These images can be used to create crater landmark maps for all craters larger than $5$ to $10$ m in diameter, which can provide a dense set of landmarks without requiring high resolution elevation maps to be created from orbit. In principle, craters can be readily detected around rovers using onboard cameras and/or lidar, then matched automatically to crater landmarks mapped from orbit, as shown in Figure \ref{fig:system_concept}. We estimate that this can enable rover absolute localization with 3-sigma error on the order of $5$ m. There are plans to fly a ShadowCam in lunar orbit that can provide images of permanently shadowed regions (PSRs) with a resolution of 1.7 m/pixel \cite{robinson2018shadowcam}, which could be used to extend such a capability to rpver absolute localization within PSRs.

This paper shows initial results of a planned three-year project - LunarNav, to develop algorithms for lunar rover absolute localization with crater landmarks. Since there are several options for sensors onboard future lunar rovers, we have developed algorithms to detect craters near rovers using three methods based on (1) 3D point clouds obtained from lidar, (2) 3D point clouds obtained from stereo vision, and (3) pattern recognition with monocular images. Section \ref{sec:related} discusses related work. Section \ref{sec:concept} describes the overall crater-based localization system concept. Section \ref{sec:data} describes data sets of real and synthetic images obtained for this work. Section \ref{sec:algo} describes the algorithms developed for the three sensing modalities. Results of quantitative performance evaluation for all three sensor modalities are shown in Section \ref{sec:eval}. These results show this approach to be very promising. Section \ref{sec:discussion} summarizes the results and outlines planned future work to complete software for a prototype crater-based navigation system.

\section{Related Work}
\label{sec:related}
There has been a great deal of work on absolute localization for vehicles on Earth \cite{campbell2020localization}; discussion here only addressed methods applicable on the surface of the Moon. Since we address missions that include very long rover traverses, we do not consider methods that rely on navigation aids from other fixed surface assets, such as radio beacons on landers. Methods that are applicable fall into several classes:

\begin{itemize}[leftmargin=25pt]
    \setlength\itemsep{0.5em}
    \item Registering onboard image or local map data to orbital images or maps
    \item Using horizon features as landmarks
    \item Celestial measurements, potentially combined with other measurements, such as accurate gravity vector measurements
    \item Radiometric ranging from lunar orbit or Earth tracking stations
\end{itemize}

Registering onboard images to orbital images has been the standard approach for human-assisted localization of Mars rovers \cite{parker2010geomorphic}. Images from a camera on the rover mast are downlinked, ortho-projected onto the ground plane, potentially mosaicked to get broader context from several images in a panorama, then correlated to an image from orbit. This can achieve accuracy on the scale of the resolution of the orbital imagery, e.g. meter-scale. This can be implemented onboard, as long as the rover carries the orbital map images. Similar concepts apply if rover images are used to first create a local digital elevation map (DEM), then this local DEM is correlated to a high resolution DEM created from orbit. Work toward onboard automatic implementation of these mehods for Mars rovers is described in \cite{gaines2020self}. For lunar applications, changes in sun angle lead to dramatic changes in terrain appearance, which makes image registration methods difficult. Sufficiently high resolution orbital DEMs are not available for as much of the Moon as monocular image coverage.

Horizon-based localization systems get attitude from celestial and/or inertial sensors and get position by matching observed horizon features to an elevation map determined from orbit; matching several such landmarks gives a set of direction vectors that can be solved for position \cite{cozman2000outdoor}. The accuracy of this method depends on the accuracy of the elevation map, on having sufficient topographic relief to generate landmarks, and on distance to those landmarks. In favorable situations, this can give errors $<$ 50m \cite{chiodini2017mars}; however, adequate terrain relief and elevation maps are not always available.

For the Moon, celestial navigation methods obtain vectors to some combination of the sun, Earth, and stars. All three degrees of freedom of attitude can be obtained this way. If the rover was on perfectly flat ground, an estimate of latitude and longitude that is compatible with this attitude estimate could give a position estimate. Since the rover will not be on flat ground in general, onboard inclinometry can measure the pitch and roll. The potential accuracy of this method depends on the accuracy of the component sensors. Simulations in \cite{yang2014simultaneous} show 1-sigma errors on the order of 100m. This is rather coarse for navigating rovers to targets.

Radiometric navigation uses radio ranging between the rover and either a lunar orbiter or Earth tracking stations to estimate rover position. In principle, this can give meter-scale position knowledge \cite{chelmins2009kalman}. The required orbiter navigation infrastructure is not currently in place to provide this service. For low latitude rovers, this would not provide a continuous update. Radiometric tracking from Earth could not provide continuous service even on the near side, due to limited availability of tracking stations; on the far side, this would not be available at all.

Of methods that use images or maps created from orbit, crater-based localization has potential to provide the greatest accuracy and to be available over the largest fraction of the surface, given that most of the surface has already been imaged by LROC-NA. This approach is more available than radiometric methods. Celestial methods need very accurate inclinometry and appears to be less accurate. Crater-based localization can be implemented with the same onboard sensing and computing resources that are needed for relative localization and for obstacle detection. These factors make it the most attractive approach we have found.

\section{System Concept}
\label{sec:concept}
As described in Section 1 and illustrated in Figure \ref{fig:system_concept}, our navigation system concept requires creating a database of crater landmarks from orbital images. This database would contain the position, diameter, and estimated depth of each crater. From crater size/frequency distributions shown in \cite{hiesinger2012old}, we expect that multiple landmarks of a situable size (i.e. diameters between 5 - 20m) would be available per 100m of traverse. Mature software for mapping craters in orbital imagery has been developed in past for orbiter localization applications \cite{cheng2003optical}.

Robotic lunar rovers will carry a sensor suite for relative localization and obstacle detection that includes wheel odometry, an IMU, either stereo cameras or a lidar, and either a sun sensor or a star camera for absolute heading measurement. We assume availability of a star camera, which is applicable for driving in sunlight and in shadow; this can give 3-axis attitude knowledge to a small fraction of a degree. With this, the relative navigation sensor suite enables deadreckoning with position error that typically can be $<$ 2\% of distance traveled. Crater detection is less expensive computationally than obstacle detection and needs to be done much less frequently than obstacle detection, so any onboard computing system that can do obstacle detection, would be able to also do crater detection.

\begin{figure*}[!t]
    \centering
    \includegraphics[width=0.95\linewidth]{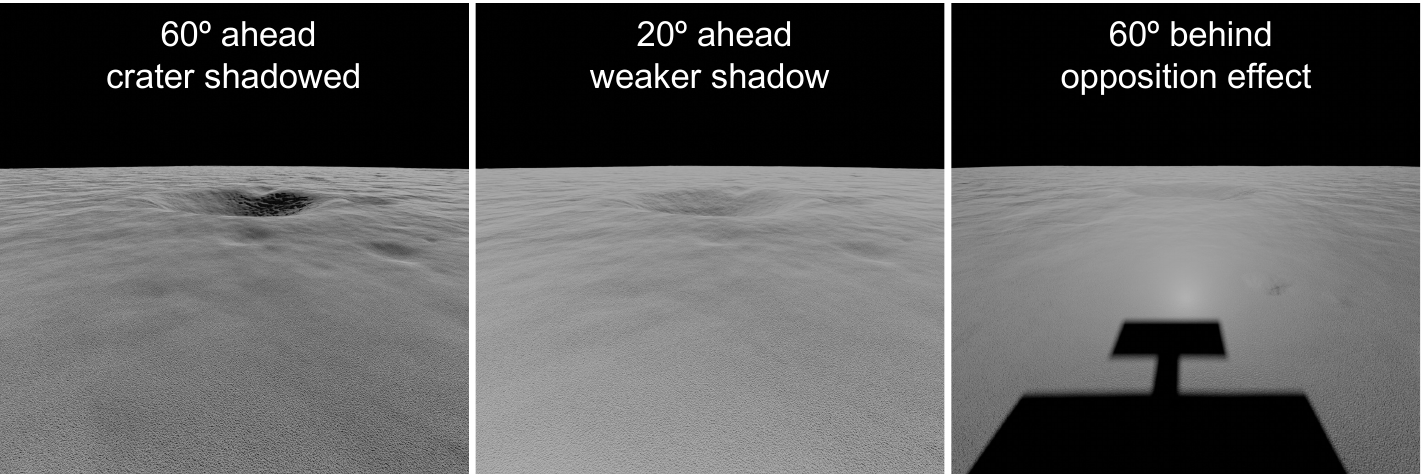}
    \caption{\textbf{Simulated images with different sun zenith angles to illustrate effect on image contrast.}}
    \label{fig:data1}
\end{figure*}

\begin{figure}[!t]
    \centering
    \includegraphics[width=0.9\linewidth]{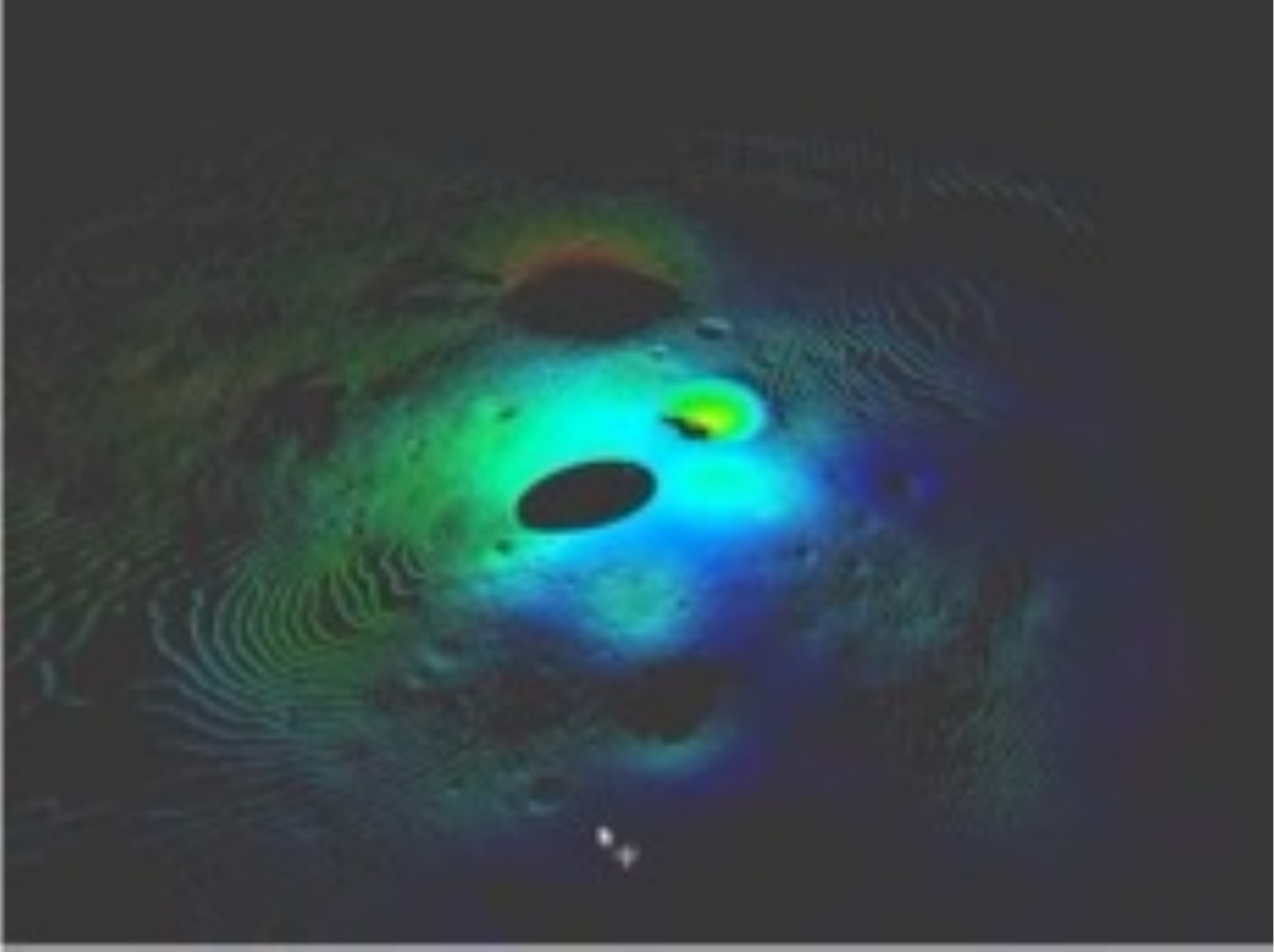}
    \caption{\textbf{Simulated LIDAR point cloud. The LIDAR is at the center of the black ellipse in the middle of the figure. Color codes elevation; blue is high, red is low. Black areas in craters are not visible to the sensor, i.e. occluded.}}
    \label{fig:data2}
\end{figure}

The relative localization system provides a prior estimate of position that at all times strongly constrains which crater(s) from the landmark database are expected to be near the rover. Craters can be detected near the rover with a combination of 3D point cloud data from stereo cameras or a lidar, image data from a camera, or reflectance image data from a lidar. This enables detecting craters with diameters roughly between 5 to 20 m whose near rims are roughly less than 20m from the rover; given the frequency of lunar craters, such craters will be encountered frequently. This detection process will not be error-free. Errors can be filtered in several ways. First, the prior position knowledge from relative localization means that nearby craters detected onboard can be compared with a small number of candidate landmarks from the database. Second, multiple nearby craters can be detected and registered to the database at any given time. Third, over short traverses (e.g. $<$ 100m), a local map of craters near the rover can be maintained, so that multiple landmarks can be matched at once. Overall, these methods should enable reliable absolute localization; given typical resolution characteristics of cameras, stereo vision, and lidar, we estimate that it should be possible to maintain a rover absolute position estimate with 3-sigma error $<$ 5m at all time. This probably is better than need by most missions, but maintaining an accurate position estimate makes future landmark recognition easier, so there is a positive feedback involved.


\section{Data Sets}
\label{sec:data}
We use both real and simulated sensor data to develop and evaluate the performance of lunar rover navigation with crater landmarks. Simulated data allows testing with larger data sets and over wider ranges of illumination conditions than are practical with real data; this enables evaluation of statistic performance metrics, which cannot be done yet with the amount of real sensor data available currently.

\begin{table}[!t]
    \centering
    \caption{\textbf{Simulated camera parameters.}}
    \renewcommand{\arraystretch}{1.5}%
    \begin{tabular}{|c|c|}
    \hline
    \textbf{Camera Parameter} & \textbf{Value} \\
    \hline
    Resolution & 1024x1024 pixels \\
    Field of View (FOV) & 90 degree \\
    Focal Length & 19mm \\
    Sensor Size & 35mm \\
    Camera Height & 1.5m \\
    Stereo Baseline & 30cm \\
    \hline
    \end{tabular}
    \label{tab:camera-param}
\end{table}

\begin{table}[!t]
    \centering
    \caption{\textbf{Simulated LIDAR parameters.}}
    \renewcommand{\arraystretch}{1.5}%
    \begin{tabular}{|c|c|}
    \hline
    \textbf{Camera Parameter} & \textbf{Value} \\
    \hline
    Vertical FOV & 75 degree \\
    Horizontal FOV & 360 degree \\
    Vertical Angular Resolution & 0.333 degree \\
    Horizontal Angular Resolution & 0.333 degree \\
    Sensor Height & 1.5m \\
    Sensor Tilt & 23 degree \\
    \hline
    \end{tabular}
    \label{tab:lidar-param}
\end{table}

\subsection{Simulated Data}
Several simulation tools were considered for this work. Based on cost and schedule considerations, we developed an initial simulation capability using the Blender computer graphics rendering engine. Future work will re-examine this choice to consider alternate scene rendering capability that is integrated with rover traverse simulation capability.

\begin{figure*}[!t]
    \centering
    \includegraphics[width=\linewidth]{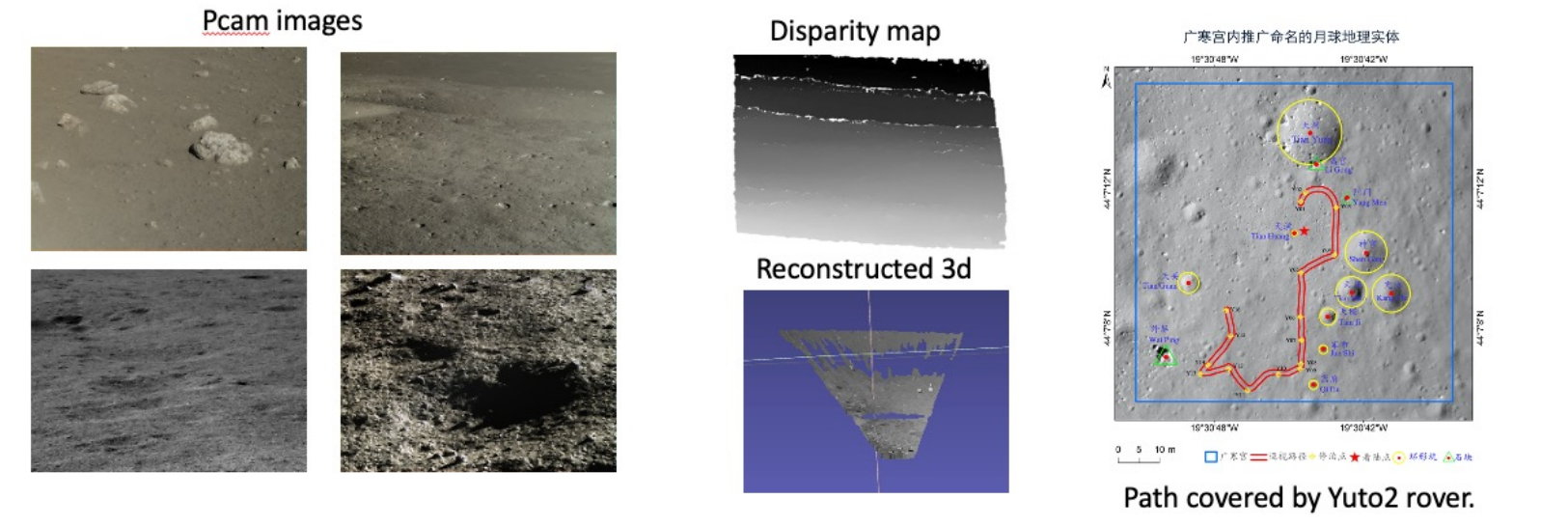}
    \caption{\textbf{Example Chang'e 3 Yutu rover images (left), an example stereo disparity map and 3D rendering (middle), and orbital image showing the rover traverse (right). Stereo disparity is inversely proportional to range; in the disparity map, bright pixels are close and dark are far.}}
    \label{fig:data3}
\end{figure*}

\begin{table*}[!hh]
    \centering
    \caption{\textbf{Yakimovsky-Cunningham camera models for Chang'e 3 PCAM. Note: Model = CAHV = Perspective, and Dimensions = 2352x1728}}
    \renewcommand{\arraystretch}{1.5}%
    \begin{tabular}{|c|c|}
    \hline
    \textbf{Left Camera} &  \textbf{Right Camera}\\
    \hline
    C = -0.000000 -0.000000 -0.000000 & C = 0.269990 -0.001864 -0.001342\\
    A = 0.000000 0.000000 1.000000 & A = -0.033442 0.001049 0.999440\\
    H = 6773.068749 0.000000 1175.500000 & H = 6729.968393 5.565861 1401.340421\\
    V = 0.000000 6791.594910 863.500000 & V = 32.980904 6792.495564 855.751098\\
    \hline
    \end{tabular}
    \label{tab:change3}
\end{table*}

\begin{table*}[!ht]
    \centering
    \caption{\textbf{Yakimovsky-Cunningham camera models for Chang'e 4 PCAM.  Note: Model = CAHV = Perspective, and Dimensions = 2352x1728}}
    \renewcommand{\arraystretch}{1.5}%
    \begin{tabular}{|c|c|}
    \hline
    \textbf{Left Camera} &  \textbf{Right Camera}\\
    \hline
    C = -0.000000 -0.000000 -0.000000 & C = 0.269990 -0.001864 -0.001342\\
    A = 0.000000 0.000000 1.000000 & A = -0.033442 0.001049 0.999440\\
    H = 6773.068749 0.000000 1175.500000 & H = 6729.968393 5.565861 1401.340421\\
    V = 0.000000 6791.594910 863.500000 & V = 32.980904 6792.495564 855.751098\\
    \hline
    \end{tabular}
    \label{tab:change4}
\end{table*}

The Blender simulation takes a lunar digital elevation map (DEM), applies a custom texture map to it, and creates a world model with a simulated sun as a light source. Within this model, a stereo camera pair is added and the simulation generates locations for the cameras and the sun. For each camera and sun location, Blender’s render engine produces a synthetic image. This image is rendered using a custom implementation of the Hapke radiometric model \cite{hapke1963theoretical,hapke1981bidirectional,hapke1993opposition} incorporated into the Blender’s path tracing algorithm to simulate accurate lighting across the surface. The simulation has multiple parameters that can be controlled, including sun position, camera extrinsic and intrinsic parameters, image size, and a few others. 

For the simulated dataset generated, images were generated for craters with diameters of 5, 7, 10, 12, 15, 17, and 20 meters, with four different approach angles for each crater to expand the size of the data set. For each crater and approach angle, the stereo camera was positioned at distances between 5 and 20 meters from the crater near rim, at 1 meter spacing; for each location, an image was rendered with sun angles of 0, 20, 40, 60, and 80 degrees from zenith. The images were rendered with 2048 x 2048 pixels and downsampled via the Lanczos interpolation procedure in the OpenCV image processing library to 1024 x 1024. The downsampling was used to avoid smoothing artifacts that were found within Blender’s renders at medium to far ranges for images rendered with 1024x1024 size directly. The final dataset contains 1,792 stereo pairs. Synthetic LIDAR images were generated similarly, by rendering 3D point clouds instead of images; sun angle was considered irrelevant for LIDAR, so that variable was ignored.

Table \ref{tab:camera-param} and Table \ref{tab:lidar-param} give sensor parameters that were simulated for stereo cameras and LIDAR. The LIDAR was simulated as a 360$^{\circ}$ scanner, because this made the most test data available with the least effort; angular and range resolution parameters are similar to commercially available sensors. Camera parameters approximate those used for stereo vision on the Curiosity and Perseverance Mars rovers. Figure \ref{fig:data1} and Figure \ref{fig:data2} show example images and a LIDAR point cloud from the simulation.

\begin{figure*}[!t]
    \centering
    \includegraphics[width=\linewidth]{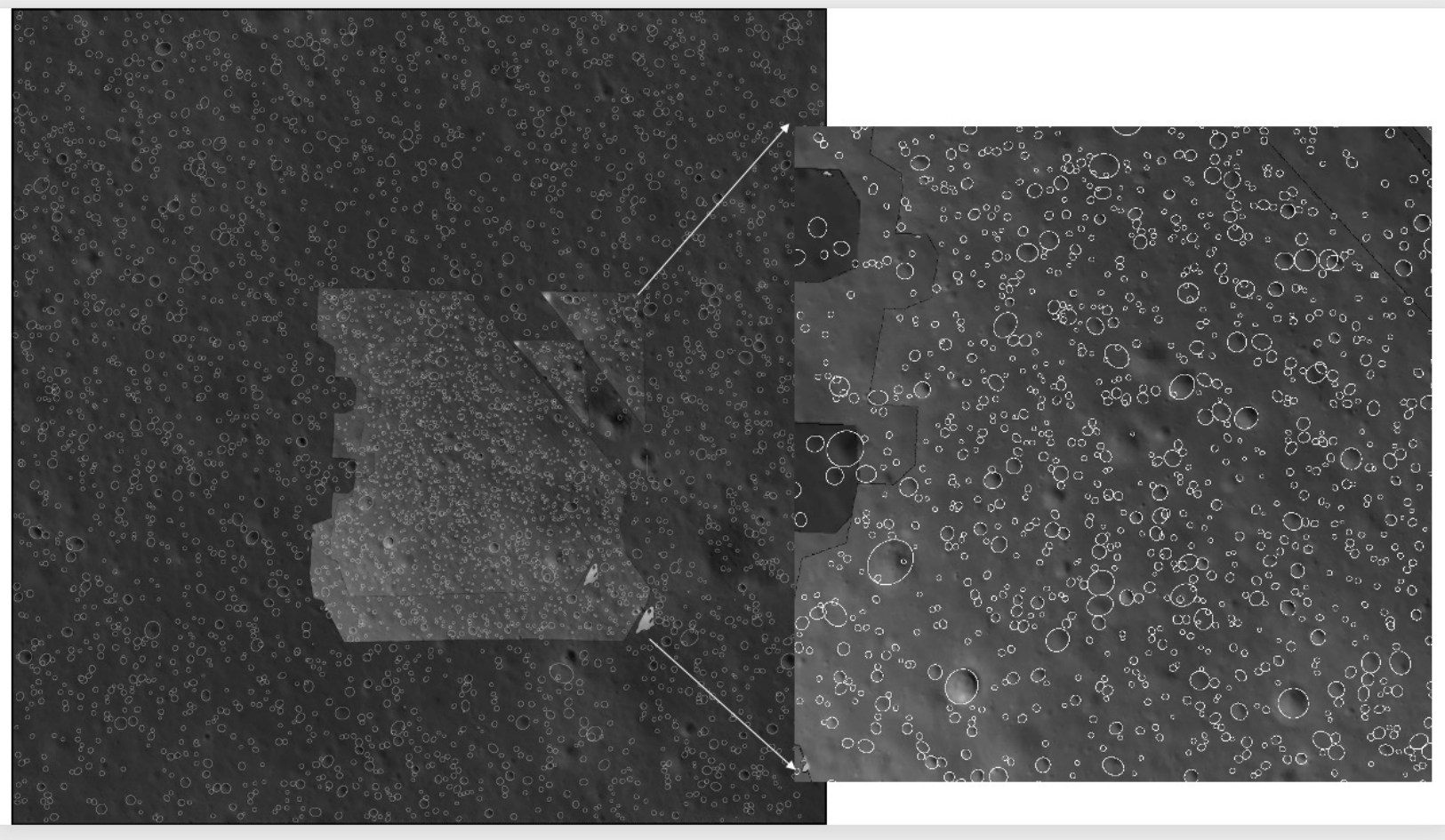}
    \caption{\textbf{Crater landmark map generation combining LRO-NAC images with higher resolution Chang’e LCAM descent imagery. Large image is from LRO-NAC; smaller overlay and the expanded view on the right is from LCAM.}}
    \label{fig:data4}
\end{figure*}

\subsection{Real Data}
For real sensor data, the original plan for to acquire a large-scale dataset of LIDAR and stereo image data from an analog site at the Cinder Lake crater field near Flagstaff, Arizona. This had to be postponed due to covid travel restrictions and forest fires; In lieu of this data, we decided to use real lunar stereo images that were available from the “panoramic” cameras (PCAM) on the Chang’e 3 and Chang’e 4 lunar rover missions, as described below. Since this is actual lunar data, in important ways it is better than the analogue stereo images that were originally planned. Ground truth labeling of craters in this imagery and registration of this imagery against orbital imagery is still in progress and hence our performance evaluation on real data was limited to qualitative results. No alternative was available for LIDAR data, so only simulated LIDAR data was used.

The Chang’e image data are publicly available\footnote{\url{https://moon.bao.ac.cn/}}. The cameras for both missions are identical, their FOV is 19.7$^{\circ}$ by 14.5$^{\circ}$, resolution is 2352x1728 pixels (color) and 1176x864 pixels (monochrome). The stereo baseline length is 27 cm. A total of 168 pairs of Chang’e 3 and 1174 pairs of Chang’e 4 PCAM images were downloaded from the site. A stereo camera self-calibration algorithm applied to this data yielded good camera models for both data sets (see Table \ref{tab:change3} and Table \ref{tab:change4}). Stereo disparity maps were computed from these images with good results, as shown in Figure \ref{fig:data3}.

We also studied usefulness of the  Chang’e lander camera (LCAM) to obtain a high resolution and high precision crater database. The LCAM image sequences (5000 images) for both missions were  downloaded and the Chang’e Entry, Descent and Landing (EDL) trajectories were recovered with terrain relative navigation (TRN) and structure from motion algorithms. Then, some of LCAM images from low altitude were ortho-rectified to a coarser resolution LRO-NAC image map (1 m/pixel) to obtain an orthoimage with resolution up to 20 cm/pixel around the lander (Figure \ref{fig:data4}). Craters were detected automatically from this high resolution map and their geographic locations, diameters, depths were extracted into a crater database. This data base will be used for rover localization algorithm development in future years. 

\section{Crater Detection Algorithms}
\label{sec:algo}
\begin{figure*}[!t]
    \centering
    \begin{subfigure}[t]{0.5\linewidth}
         \centering
         \includegraphics[width=\textwidth]{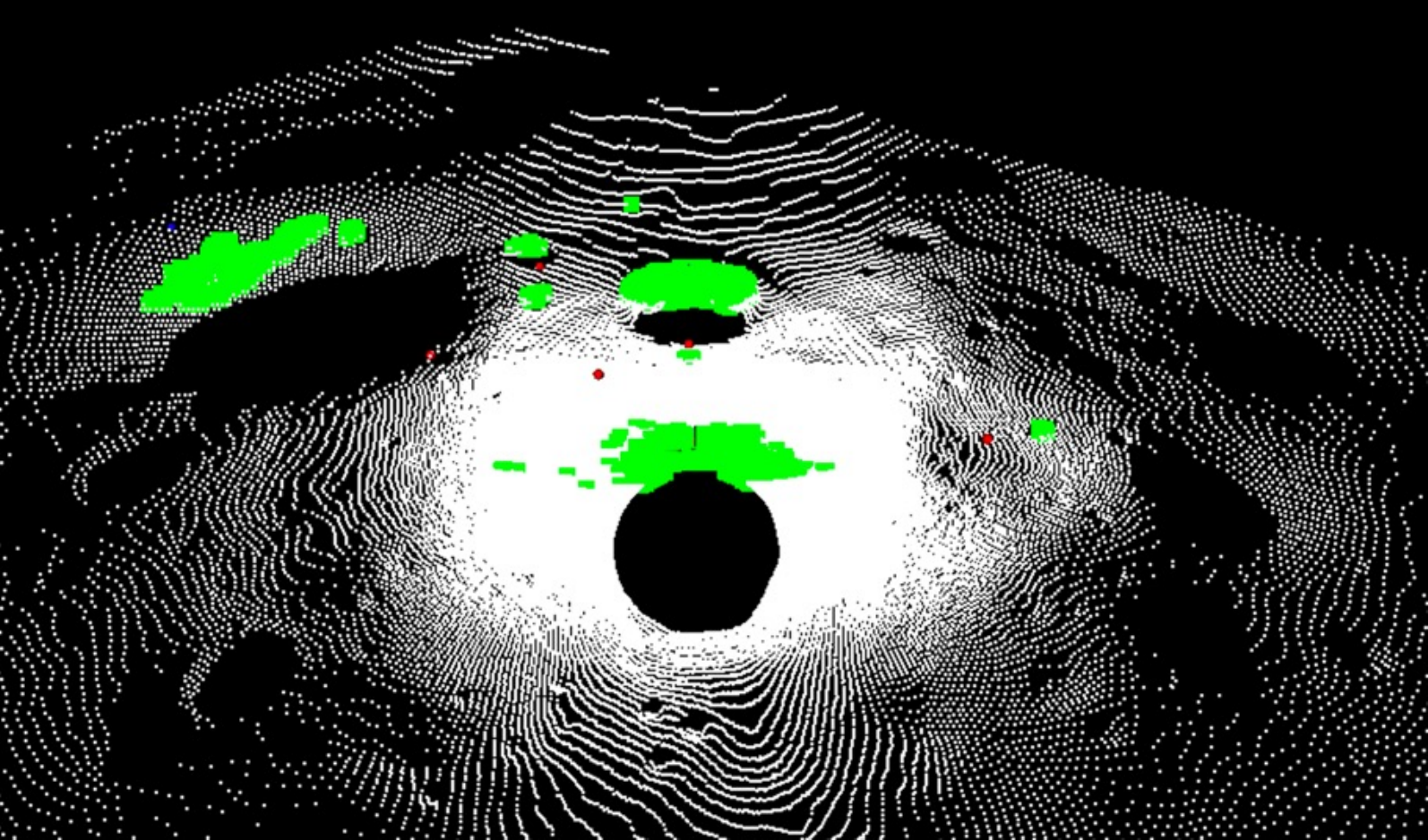}
         \caption{Using surface normal to find candidate crater back walls. Locations in a point cloud where the surface normal points toward the sensor are highlighted in green. The sensor was located in the middle of black circle with no data in the lower middle of the point cloud.}
         \label{fig:lidar-algo1}
    \end{subfigure}
    \hspace{5mm}
    \begin{subfigure}[t]{0.41\linewidth}
         \centering
         \includegraphics[width=\textwidth]{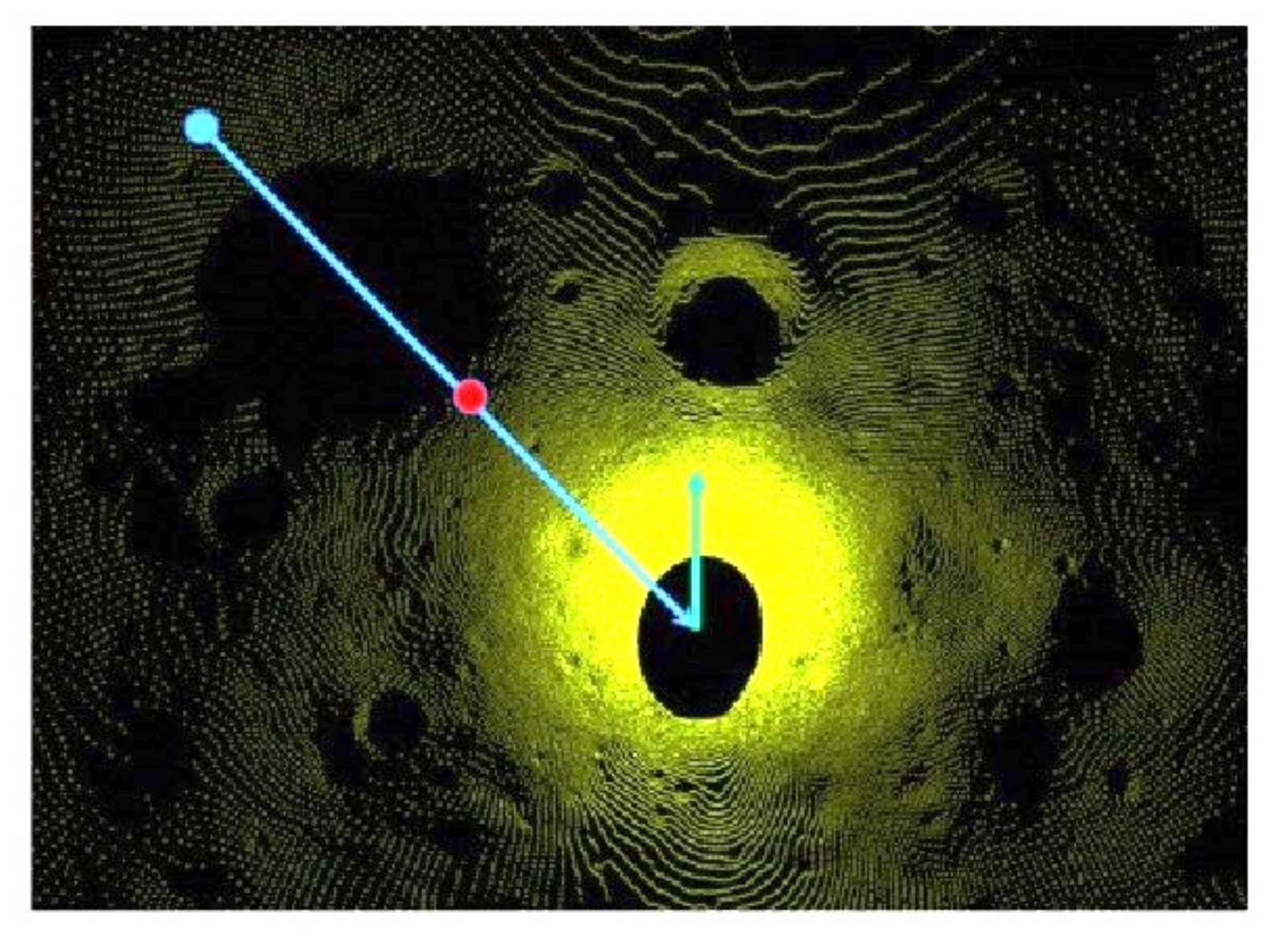}
         \caption{Obtaining initial estimate of crater center. Top-down view of point cloud; robot point and viewing direction is shown by green arrow. Blue dot is the crater back wall cluster centroid; red dot is the detected front rim.}
         \label{fig:lidar-algo2}
    \end{subfigure}\\
    \vspace{5mm}
    \begin{subfigure}[t]{0.5\linewidth}
         \centering
         \includegraphics[width=\textwidth]{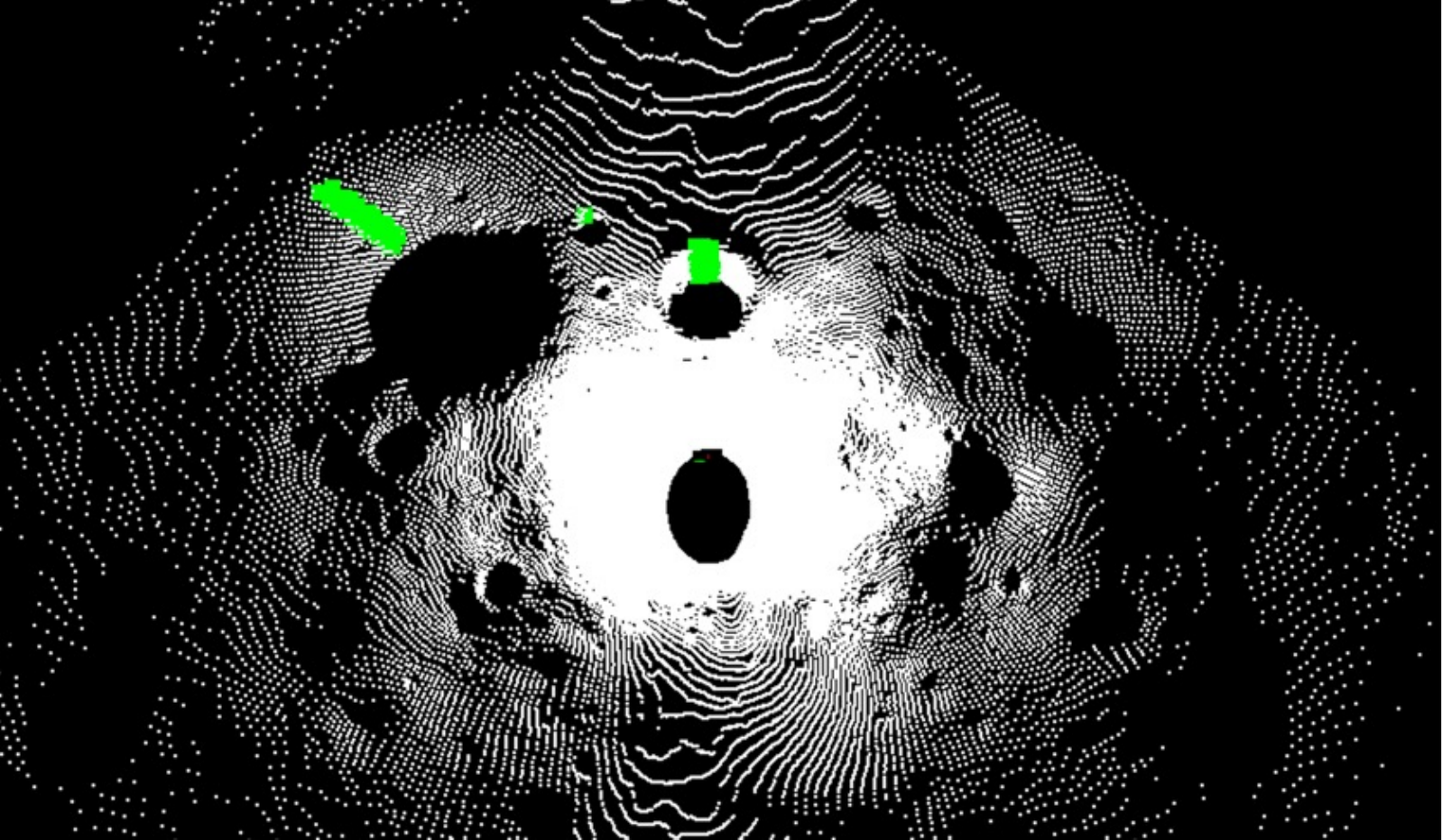}
         \caption{Filtering surface normals to refine crater center estimate. The most distant point on the green regions is taken as the far rim of the crater.}
         \label{fig:lidar-algo3}
    \end{subfigure}
    \hspace{5mm}
    \begin{subfigure}[t]{0.45\linewidth}
         \centering
         \includegraphics[width=\textwidth]{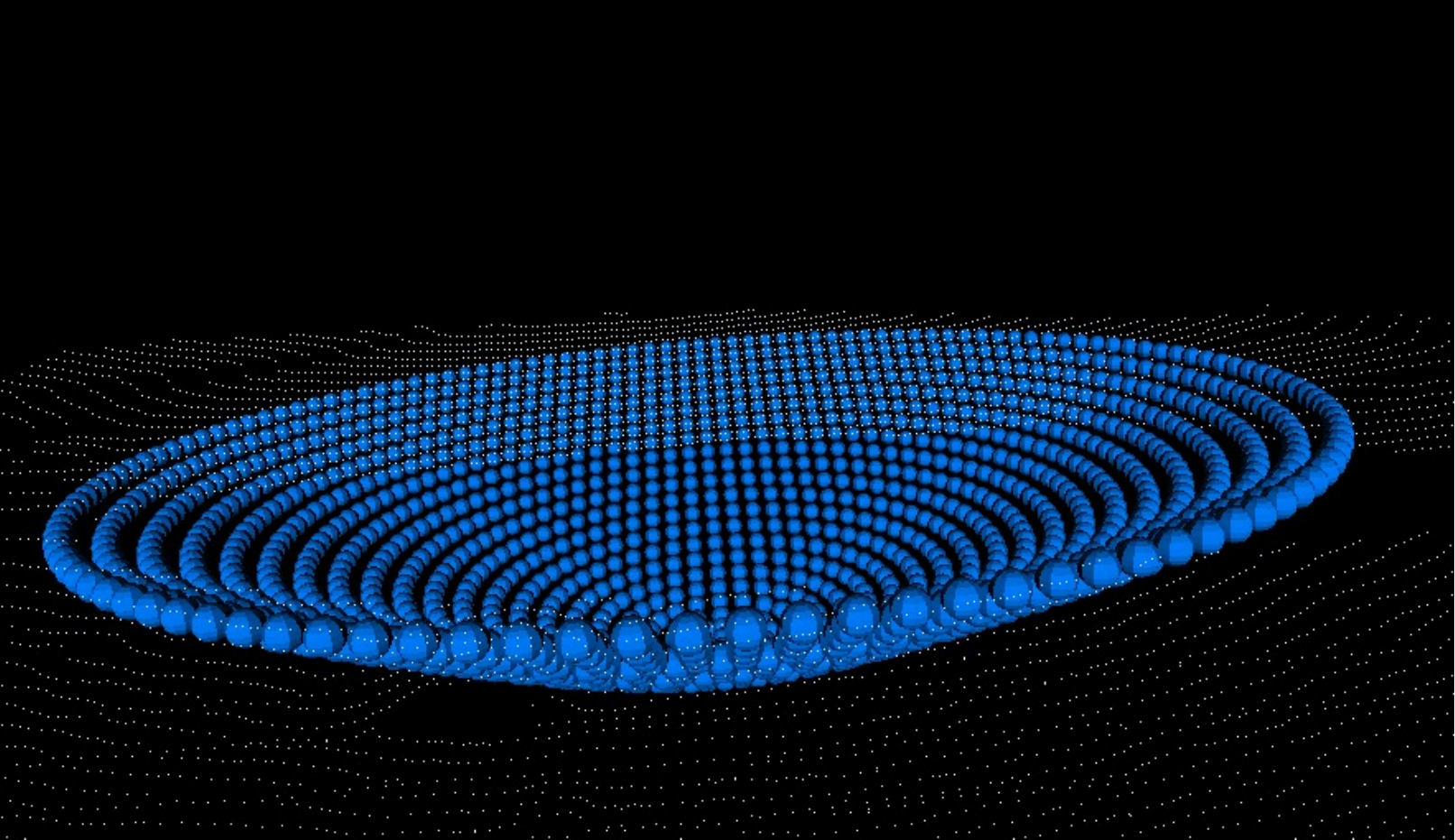}
         \caption{Subsampled parametric crater points overlayed on input point cloud.}
         \label{fig:lidar-algo4}
    \end{subfigure}\\
    \vspace{5mm}
    \begin{subfigure}[t]{0.5\linewidth}
         \centering
         \includegraphics[width=\textwidth]{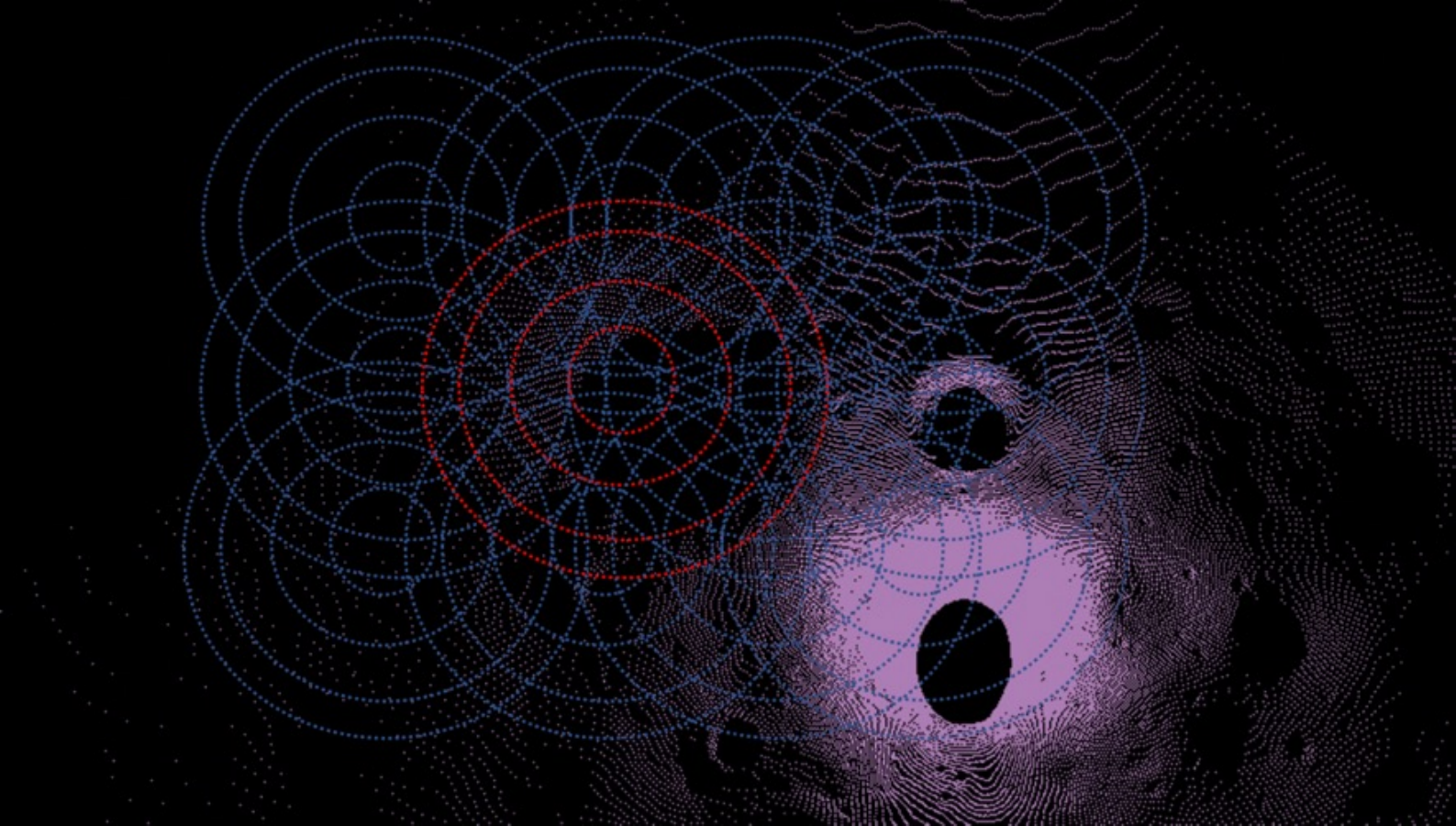}
         \caption{Top-down visualization of parametric model overlayed at different locations around the estimated crater centroid to match with. When the model’s location highlighted in red shows the location with the highest match score.}
         \label{fig:lidar-algo5}
    \end{subfigure}
    \hspace{5mm}
    \begin{subfigure}[t]{0.45\linewidth}
         \centering
         \includegraphics[width=\textwidth]{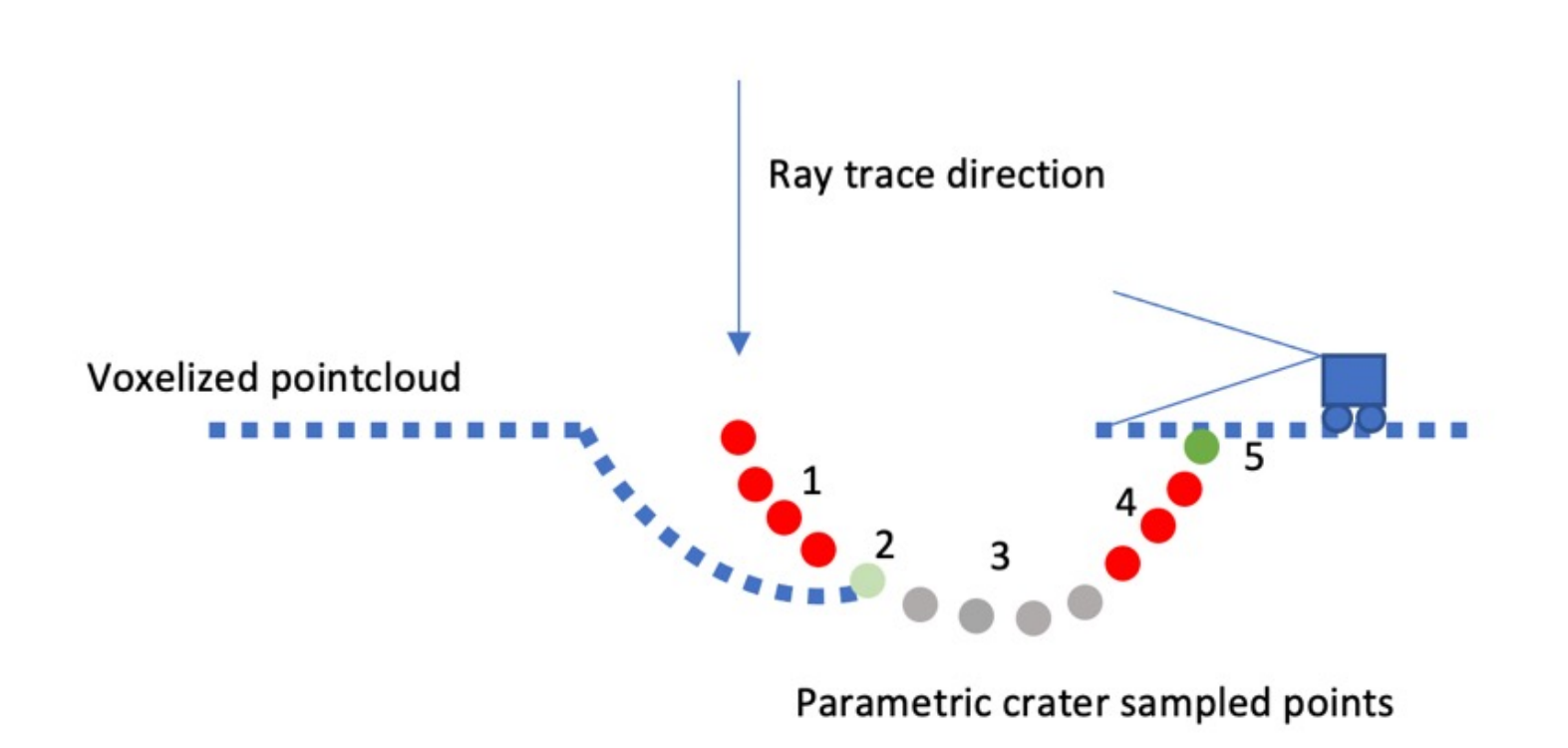}
         \caption{Side view of matching parametric crater model with voxelized point cloud.}
         \label{fig:lidar-algo6}
    \end{subfigure}
    \caption{\textbf{LIDAR-based Crater Detection Algorithm}}
    \label{fig:lidar-algo}
\end{figure*}

Planetary rovers to date have all used stereo cameras for terrain imaging and 3D perception \cite{goldberg2002stereo}. Future rovers (to Moon and Mars) might have LIDAR, either by itself or in addition to one or more cameras. Thus, any combination of these sensors could be used for crater detection. To cover this set of options and to create a foundation for identifying the best approach in the future, crater detection algorithms were developed for three classes of technical approach: (1) Using 3D point clouds from LIDAR, (2) Using 3D point clouds from stereo vision, and (3) Using deep-learning based pattern recognition with monocular images.

The original plan was to treat crater detection as a process done independently from any knowledge of the crater landmark map and rover position. Work this year showed that more reliable crater detection results can be achieved by assuming approximate prior knowledge of rover position, which is realistic in practice, and using that to allow the crater detection process to invoke 3D models of craters expected to be around the rover based on this prior knowledge. Such approximate prior knowledge has been used for the LIDAR-based approach developed to date, but has not yet been carried over to the other approaches. Geometric analysis methods have been appointed to the point clouds from LIDAR and stereo vision; machine learning with neural nets has been used for monocular images. Alternatives or combinations of methods may be explored in the future.

\subsection{LIDAR-based Approach}
The LIDAR pointcloud detection algorithm takes in the following inputs:
\begin{itemize}[leftmargin=25pt]
    \setlength\itemsep{0.5em}
    \item LIDAR point cloud acquired by a rover
    \item Crater diameter and depth parameters expected to be found in the point cloud, based on the orbital map and approximate rover position.
\end{itemize}

\subsubsection{Stage 1: Detecting potential crater locations} 
The algorithm begins by fitting a ground plane to the point cloud and rotating it so that the Z axis is aligned with the gravity vector. Then it computes the surface normals for a patch around each point in the point cloud and filters out those normal that do not point towards the sensor. The green colored points in Figure \ref{fig:lidar-algo1} show the accepted point normals as a result of this process. This step is meant to detect potential locations of crater back walls. These regions are clustered and filtered for size to provide regions that are candidates to be crater back walls. The center of each cluster is used in the next stage described below.

\subsubsection{Stage 2: Estimating detected crater diameter} 
To obtain initial estimates of the diameter of detected craters, the algorithm must first estimate the location of the back and front rims. First, the algorithm computes the centroids of the clusters. An example is shown by the blue ball in Figure \ref{fig:lidar-algo2}. The angle and distance of this centroid with respect to the origin is used to repeatedly ray trace the voxelized point cloud in order to find the front edge of the crater. The raytracing is done starting at a point above the point cloud and pointing towards the ground. The front edge is determined to be the point of transition between no-data to data points along this path as shown by the red dot in Figure \ref{fig:lidar-algo2}. The mid-point between these two locations is a rough estimate of the crater center, which must be refined because the blue dot is not on the rim of the far side of the crater. Detecting the back rim is done by re-filtering the point normals in each cluster. Only normals that point toward the initially estimated crater center are kept. Then we find the furthest point along the direction towards the cluster centroid (Figure \ref{fig:lidar-algo3}). The distance between this detected back rim point and the front rim point yields an estimate of the diameter of the detected craters as well as an updated estimate of the crater center.

\subsubsection{Stage 3: Matching detected craters with input craters}
Once initial estimates of detected craters have been found, we attempt to correspond the input crater candidates (from the orbital map) to the detected craters in the point cloud. For each detected crater, the algorithm finds all input craters that have a similar diameter. Given a detected crater and an input crater of similar diameter, we now compute a score of how well the input crater would match with the detected crater. The matching process starts by generating a parametric model of the input crater using the given diameter and depth. Points from the model are sampled in a uniform fashion as shown in Figure \ref{fig:lidar-algo4}. The algorithm then places this model at locations in a 2D grid centered around the estimated centroid of the detected crater as shown in Figure \ref{fig:lidar-algo5}. An octree is generated from input point cloud in order to match this model to a voxelized representation of the detected crater. A score is computed to see how well the model fits with the detected crater voxels when placed at these locations. The location with the highest score is kept for further ranking to compare with other input craters. 

The score that is computed is the sum of individual scores on each point in the parametric model as follows (see Figure \ref{fig:lidar-algo6}):
\begin{itemize}[leftmargin=25pt]
    \setlength\itemsep{0.5em}
    \item For points that are near the front edge of the crater before the front rim, reduce the score if there are occupied voxels that are nearby. These are shown in area 4 in the diagram. The reason for this is that we want to capture the fact that this area typically should see no LIDAR returns due to occlusions.
    \item For rim points of the parametric crater, reduce the score if there are no occupied voxels nearby. If there are occupied voxels, then make sure they are close enough, otherwise penalize the score. The point shown in area 5 is not penalized because it is close enough
    \item For other points in the parametric crater, if we found a voxel that is higher than the point, penalize the score. These are area 1 points. Area 2 points are not penalized because it is close enough.
    \item For all other points in the parametric crater, do not penalize if there are no voxels matching. These are area 3 points.
\end{itemize}

Finally, the algorithm returns the locations of the input craters detected in the point cloud. This fitting method should improve estimates because it uses all of the 3D points within the crater, not just those along one line through the center; it also enables rejecting candidate crater detections that are inconsistent with all nearby craters in landmark database.


\begin{figure*}[!t]
    \centering
    \includegraphics[width=0.8\linewidth]{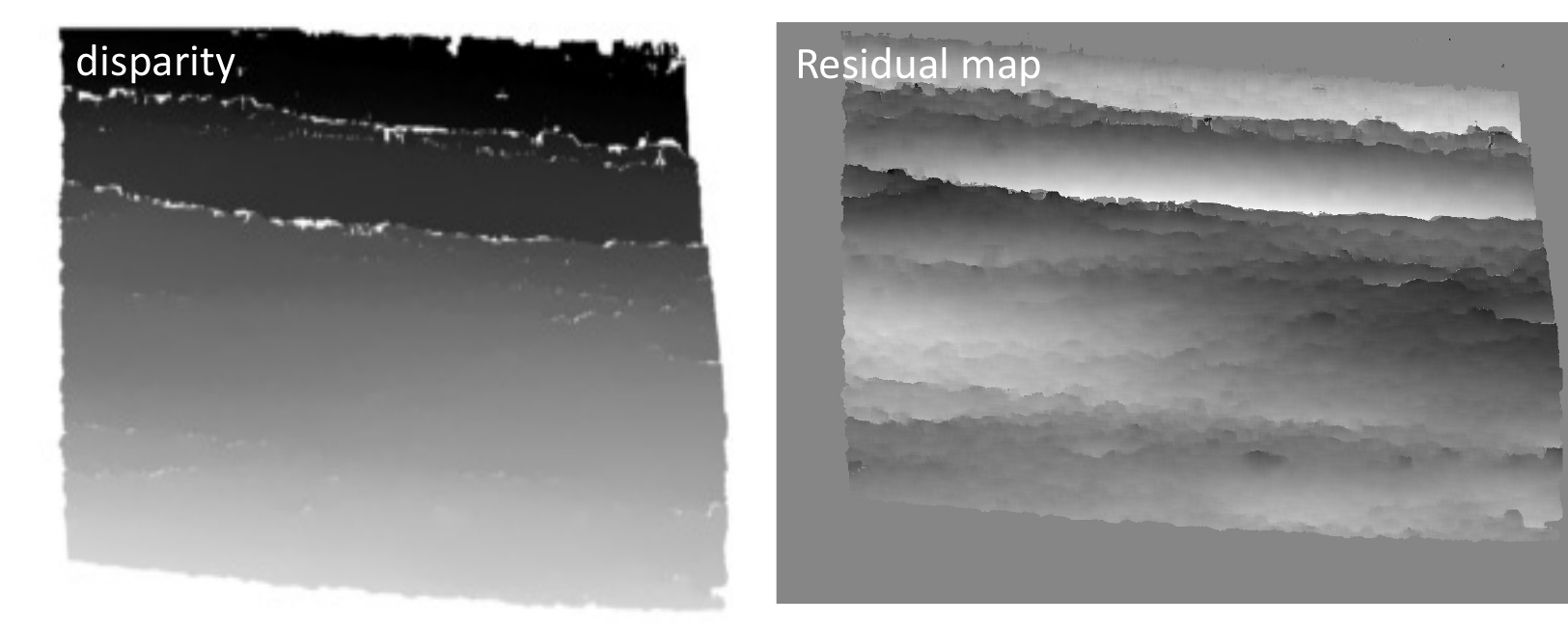}
    \caption{\textbf{A residual disparity map is computed by differencing the raw disparity map and a plane fit of the raw disparity map. See text for discussion.}}
    \label{fig:stereo-algo1}
\end{figure*}

\begin{figure*}[!t]
    \centering
    \includegraphics[width=0.8\linewidth]{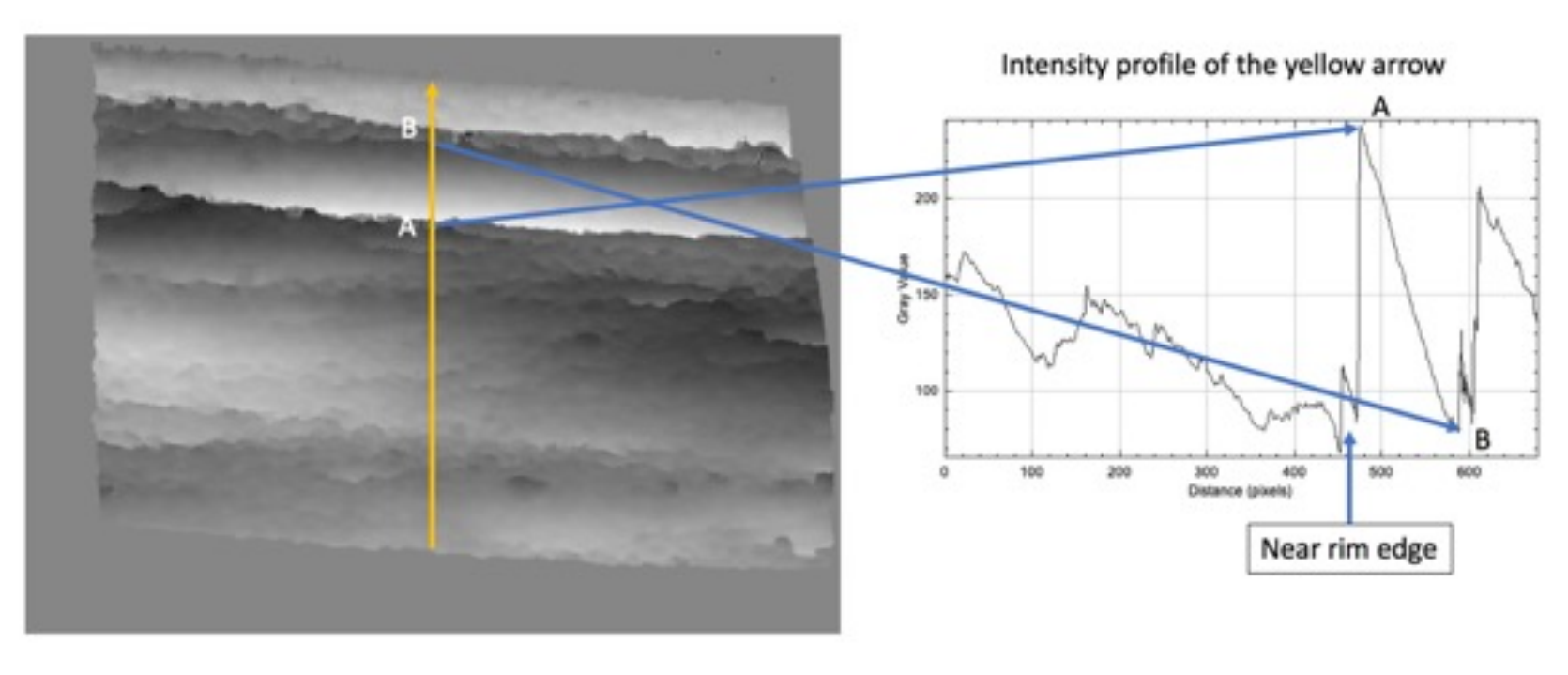}
    \caption{\textbf{Vertical profiles of the residual disparity map highlight the range jump at the near rim and the approximately linear slope on the far side of the crater interior. These cues are used for detection.}}
    \label{fig:stereo-algo2}
\end{figure*}

\begin{figure*}[!t]
    \centering
    \includegraphics[width=0.95\linewidth]{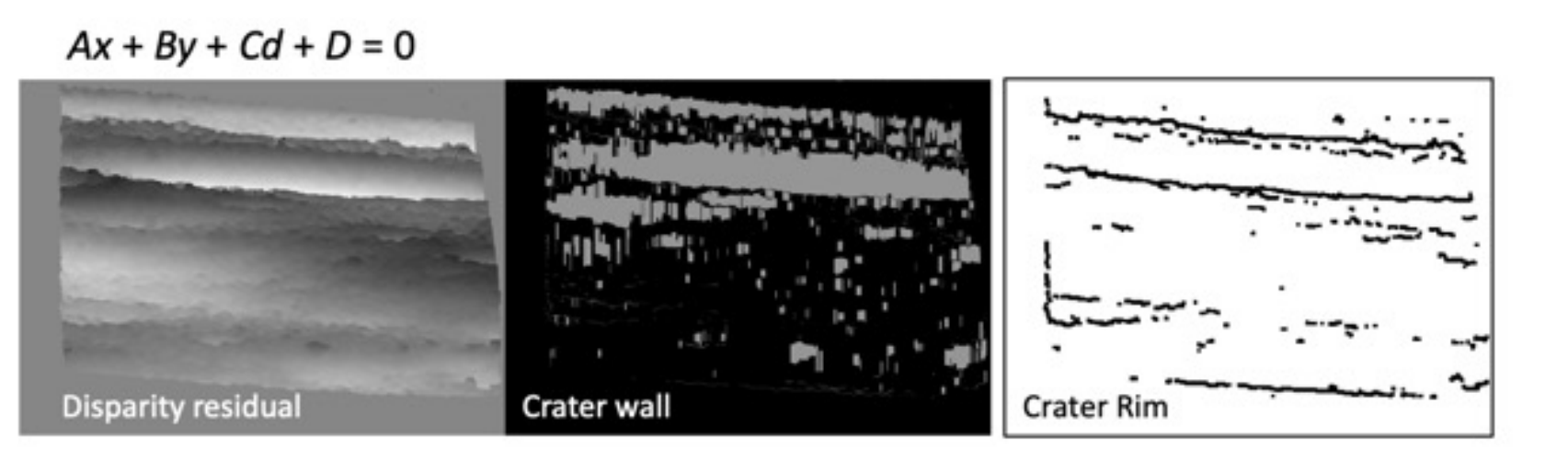}
    \caption{\textbf{Candidate back wall regions and rim contours are paired to detect the crater.}}
    \label{fig:stereo-algo3}
\end{figure*}

\begin{figure}[!t]
    \centering
    \includegraphics[width=0.8\linewidth]{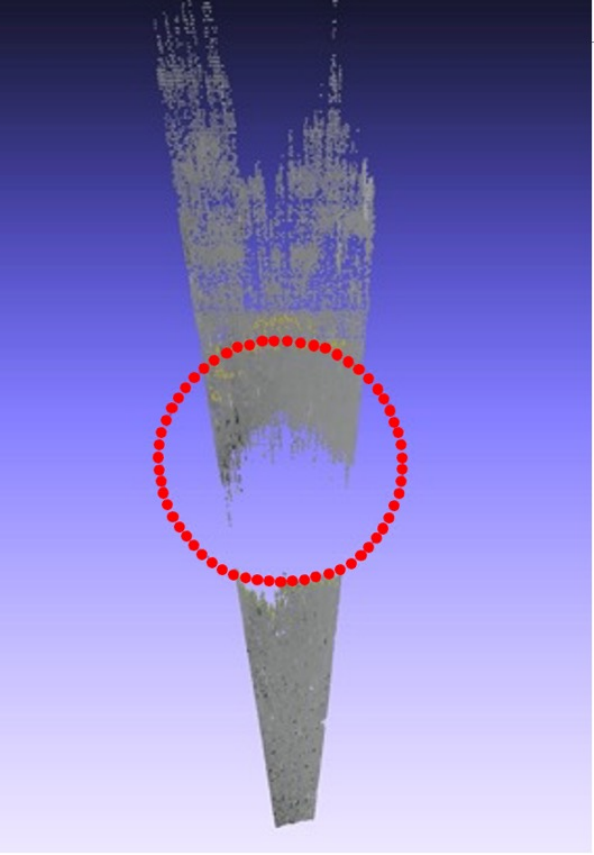}
    \caption{\textbf{Vertical profiles of the residual disparity map highlight the range jump at the near rim and the approximately linear slope on the far side of the crater interior. These cues are used for detection.}}
    \label{fig:stereo-algo4}
\end{figure}

\subsection{Stereo Image-based Approach}
Standard stereo matching algorithms are applied to the stereo image pairs to compute stereo disparity maps, from which 3D point clouds are generated. The range resolution of stereo vision degrades quadratically with true range, so 3D point clouds from stereo inherently are noiser than LIDAR at large range. There also tends to be more noise at range discontinuities, e.g. the near and far rims of craters. Initial effort to apply the crater detection method developed for LIDAR to 3D point clouds from stereo did not produce satisfactory results. Therefore, an alternate method for processing the stereo data was developed, as described below. This does not yet include a parametric model-matching stage using approximate prior knowledge of rover position to index predicted craters from the landmark database. Thus, results (as shown in Section \ref{}) so far with stereo are noisier than with LIDAR, though still promising for rover localization.

Stereo image-based crater detection approach starts with the disparity map; note that stereo disparity at a pixel is inversely proportional to the range to the surface at that pixel. First, we fit the disparity map with a plane as:
\begin{equation}
    Ax + By + Cd + D = 0
\end{equation}
where $x$, $y$ index image row and column, $d$ is disparity, and $(A,B,C,D)$ are parameters of the plane fit. The purpose of this plane fitting is twofold. First, the fitted plane defines the general surface plane of the imaged region. Because the rover is not always level, and it could be on a slope, this fitted plane limits any confusion related to the attitude uncertainty. Second, the derivation from the fitted plane helps reveal the crater topography.

Two primary cues are used for crater detection (Figure \ref{fig:stereo-algo1}). First, the near side rim of craters show a sharp discontinuity in range. Second, the far side of the crater interior walls show a smooth, positive slope. In Figure \ref{fig:stereo-algo2}, the residual between the fitted disparity plane and the disparity map is plotted against image row, going up the column. The sharp jump in the profile happens at the near rim of the crater, and the roughly linear slope following it is the far side of the crater.

The crater detection algorithm finds line segments in each column of the residual disparity map that could be the far side of a crater. Adjacent columns with candidate crater regions form connected image regions that are found with a standard connected component labeling algorithm (Figure \ref{fig:stereo-algo3}, center), then small regions are deleted as noise. Candidate crater rims are found by detecting range discontinuities (Figure \ref{fig:stereo-algo3}, right). Pairing candidate rims and far side regions and filtering out weak candidates yields crater detections. Simple fitting algorithms are then used to estimate the crater center position and diameter (Figure \ref{fig:stereo-algo4}).

\subsection{Monocular Image-based Approach}

\begin{figure*}[!t]
    \centering
    \includegraphics[width=\linewidth]{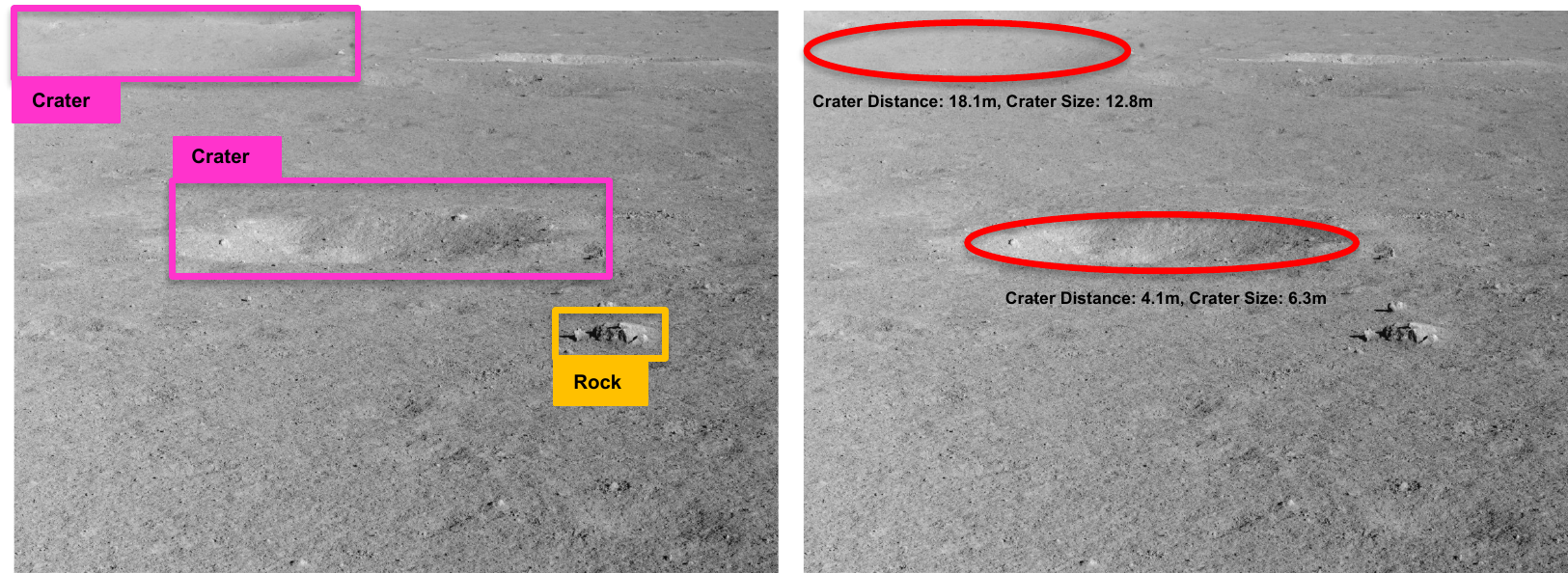}
    \caption{\textbf{Crater detection algorithm using appearance in monocular images. (Left) A CNN based classifier is used to find candidate craters. (Right) Next, model-based ellipse fitting is used to estimate crater size and position for all the craters}}
    \label{fig:mono-algo1}
\end{figure*}

\begin{figure}[!t]
    \centering
    \includegraphics[width=\linewidth]{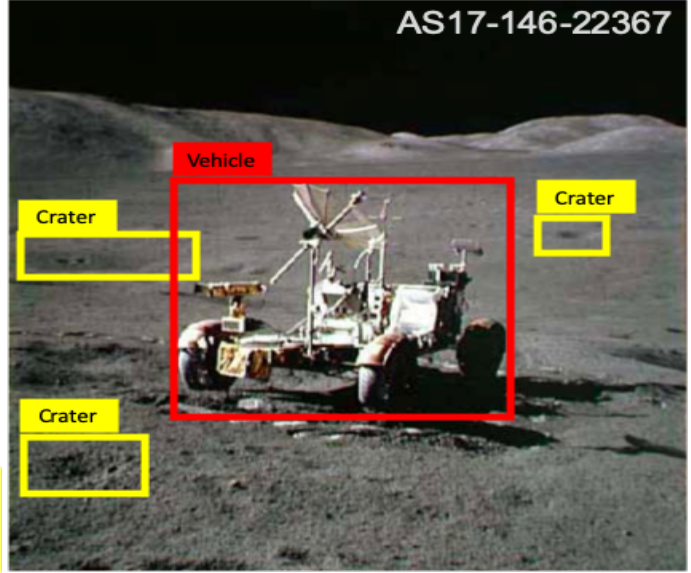}
    \caption{\textbf{Example training Image from Apollo mission with Ground Truth labels}}
    \label{fig:mono-algo2}
\end{figure}

The shading pattern of craters potentially allows them to be detected directly from image appearance, without first using 3D point cloud information for detection based on 3D shape cues. Appearance-based detection may enable detection at greater range and may also allow more reliable crater detection when paired with 3D point cloud analysis. Appearance-based detection is essentially a pattern recognition problem, for which machine learning algorithms based on convolutional neural networks (CNN) currently achieve the state of the art performance \cite{zhao2019object}. CNN-based crater detection has shown promise when applied to orbital images \cite{downes2020deep}, but has not yet been applied to rover images. In this work, we develop such an approach.

Our method follows a two-step process: First, a data-driven classifier based on a CNN model is used to find likely craters (and other potential objects of interest such as rocks, etc.) as rectangular patches in image space. Next, crater parameters such as diameter and position are estimated for all the candidate craters, using model-based image processing techniques such as ellipse fitting, as shown in Figure \ref{fig:mono-algo1}. In particular, our CNN architecture is based on the Yolov4 \cite{bochkovskiy2020yolov4}, a state-of-the-art real-time object detection model. The model takes as input monocular images and outputs bounding box predictions around each of the objects detected in the image, labels the detected classification, and has a corresponding confidence score with each label. The CNN model was implemented in Tensorflow, initialized using weights from \cite{lin2014microsoft}, and trained on a hand-labeled dataset of ~1000 lunar images from the Apollo mission using supervised learning.

Figure \ref{fig:mono-algo2} shows an example image that was used for training the CNN model. The hand-labeled dataset consisted of three object classes – craters, rocks and artifacts. Here, “artifact” represents any human-made object, such as landers, etc. The model achieved an overall accuracy of 94\% on a test set of lunar Apollo images. Furthermore, we also qualitatively evaluated the model’s capacity to transfer learn to a different dataset using lunar images from the Chang’e 3 and Chang’e 4 missions. 

For crater parameter estimation, an image patch corresponding to each bounding box detection is used as input; pre-processing techniques are then used to improve the contrast of the image. Following this, the contours of the craters are found and convex hulls are found for these contours, which enables completion of circles for partially detected crater rims. Finally, the convex hulls are then fit with ellipses and their size and position is estimated using stereo range data.

\section{Performance Evaluation}
\label{sec:eval}
We evaluated the performance for the different crater detection algorithm described in Section \ref{sec:algo}. For each approach, we conducted qualitative evaluation on a few test cases and quantitative evaluation of statistical performance metrics. Where real data was available (stereo and monocular image approaches), this was used for qualitative evaluation. Quantitative evaluation requires large data sets, including that a variety of solar illumination angles for the stereo and monocular image-based approaches. It was prohibitively difficult to obtain large sets of real data for statistical evaluation, so quantitative evaluation was done only with simulated data, using the data sets described in Section \ref{sec:data}.

\subsection{Performance Metrics}
Thorough performance evaluation will be a function of many parameters, and will depend fundamentally on the ability to detect individual craters and to estimate their positions and diameters. This depends on (1) crater size and distance from the rover, (2) rover camera/lidar sensor parameters, including angular resolution, range resolution, field of view, and sensor height above the ground, (3) lighting conditions, and (4) other characteristics of terrain geometry, like slope. For this document, we distill a few key performance parameters (KPPs) and anticipate that craters with diameters $\geq$ 5 m will be readily detectable from distances of at least 15 m, in many but not necessarily all lighting conditions. For example, detection probability ($P_d$) of 0.5 for 5 m craters at 15 m range should be a conservative estimate. With a notional stereo camera system with angular resolution of 1 milliradian/pixel and binocular camera baseline of 30 cm, 3$\sigma$ errors in estimating the position and diameter of such craters should be $<$ 2 m each.

For LIDAR and stereo image-based approaches, we evaluated two KPPs: the probability ($P_{d}$) of correctly detecting crater landmarks that were searched for in the point clouds and the standard deviation ($\sigma_{p}$) of the error in estimating crater position relative to the rover. Since the LIDAR-based algorithm already uses approximate rover position knowledge to search for candidate crater landmarks from the landmark database, evaluating error in estimating crater diameter is not meaningful and was omitted from the entire evaluation. For the monocular image-based approach, crater detections is done only in image space, but not crater position or size information in 3D; therefore evaluation of the monocular method addressed only $P_{d}$.

\begin{figure*}[!t]
    \centering
    \begin{tabular}{cccc}
    \includegraphics[width=0.24\linewidth]{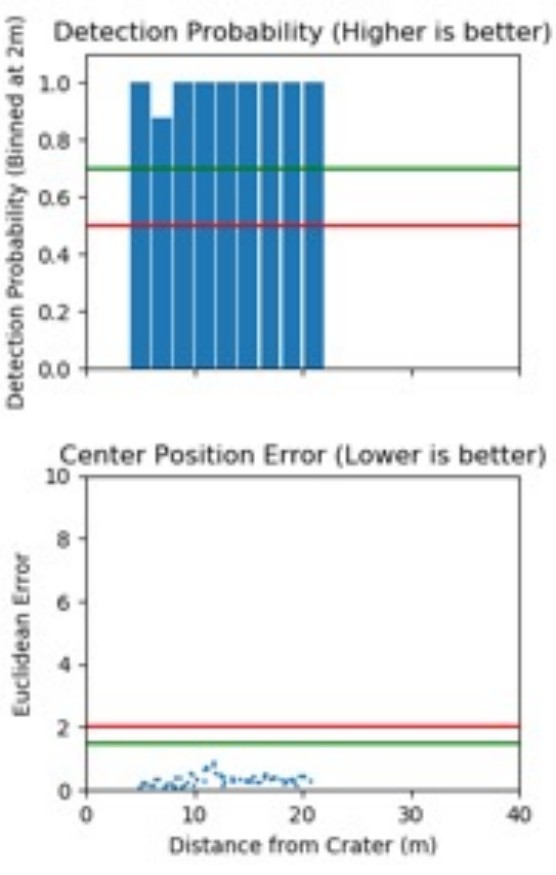} &
    \includegraphics[width=0.24\linewidth]{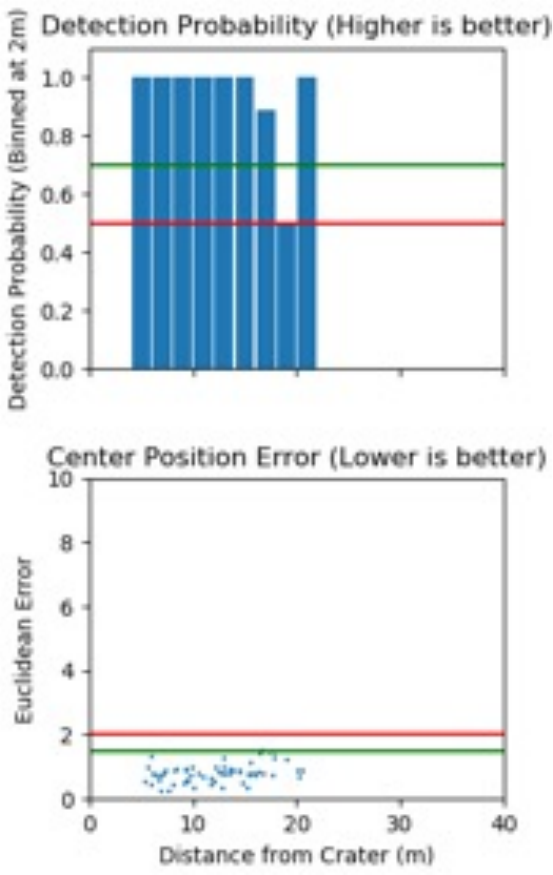} &
    \includegraphics[width=0.24\linewidth]{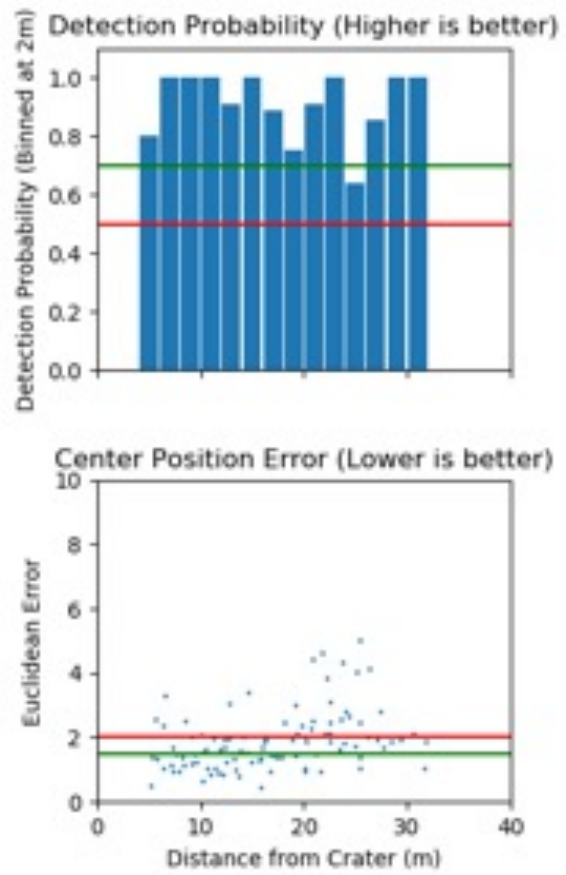} &
    \includegraphics[width=0.24\linewidth]{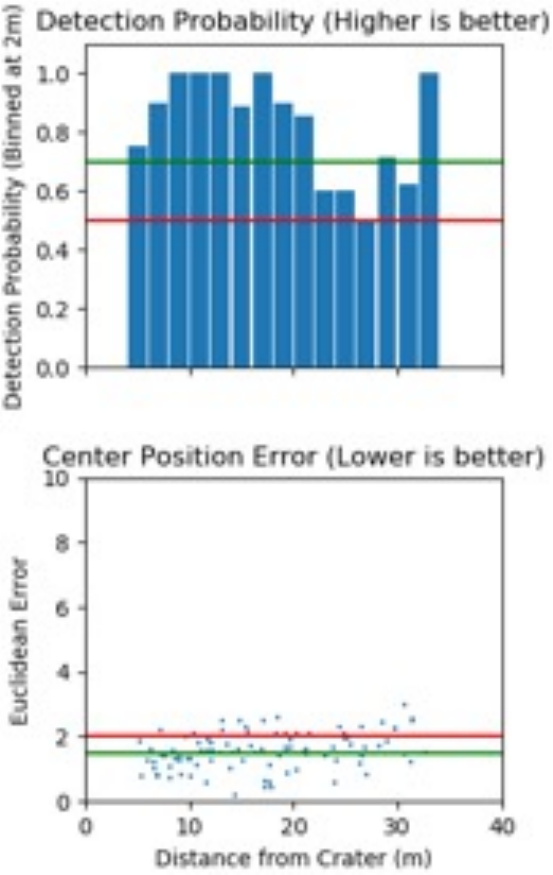} \\
    \end{tabular}
    \caption{\textbf{Results of LIDAR-based crater detection on simulated data set. Red and green lines show threshold and goal KPP levels, respectively.}}
    \label{fig:lidar-eval}
\end{figure*}

\begin{table}[!t]
    \centering
    \caption{\textbf{LIDAR-based crater position estimation error.}}
    \renewcommand{\arraystretch}{1.5}%
    \begin{tabular}{|c|c|}
    \hline
    \textbf{Crater Diameter (m)} & \textbf{$3\sigma$ Position Error (m)} \\
    \hline
    5 & 0.28 \\
    10 & 1.04 \\
    15 & 1.77 \\
    20 & 1.94 \\
    \hline
    \end{tabular}
    \label{tab:lidar-table}
\end{table}

\subsection{Performance Evaluation for LIDAR Approach}
Figure \ref{fig:lidar-eval} shows plots of $P_d$ and crater center position error as a function of the distance to the near rim of the crater. Detection performance met our objectives to the maximum range tested. Table \ref{tab:lidar-table} shows $\sigma_{p}$ as a function of crater diameter, computed for all craters between 15 and 20 meters from the rover, since the KPP for this parameter was defined for craters 15 m away. KPP goal levels were achieved for both parameters. Performance tends to be worse with increasing crater diameter, for two reasons. First, there is greater difficulty in precisely and consistently hand-annotating the ground truth crater rim for larger craters; second, the parametric matching process becomes less precise as the size of the crater grows due to multiple similar scoring candidates within the vicinity of the actual ground truth center. Finally, we see more variation in error with the 15 m diameter crater due to its shallower depth, resulting in less precision in detecting the crater rim edges.

\begin{figure*}[!t]
    \centering
    \includegraphics[width=\linewidth]{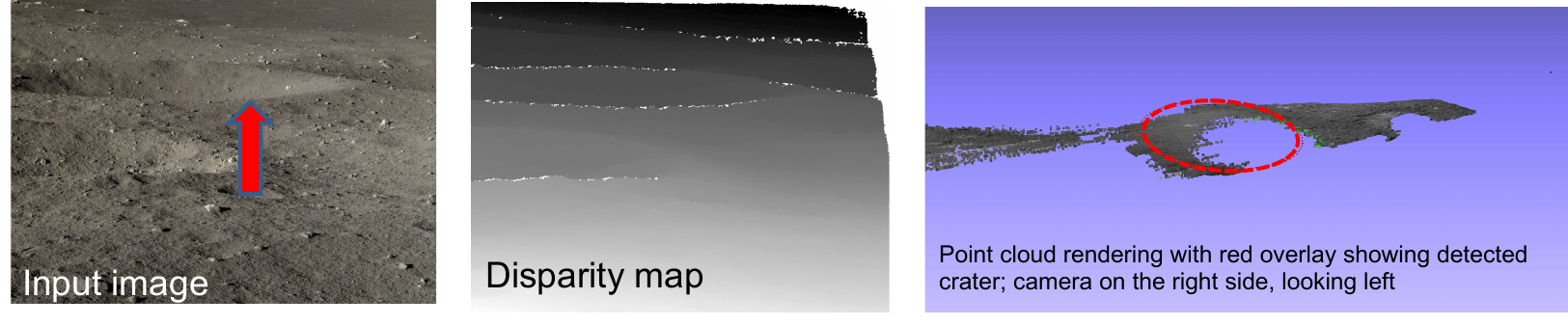}
    \caption{Qualitative stereo-based crater detection result with Chang’e rover imagery. The red arrow marks an 8 m diameter crater with near rim 24 m from the rover. In the point cloud rendering, missing data in the center of the crater and on the right side of the rendering corresponds to crater interiors that are not visible from the camera. Missing data on the left side of the rendering corresponds to lower terrain beyond the crater, which is occluded by the far rim of the crater.}
    \label{fig:stereo-qual1}
\end{figure*}

\begin{figure*}[!t]
    \centering
    \includegraphics[width=\linewidth]{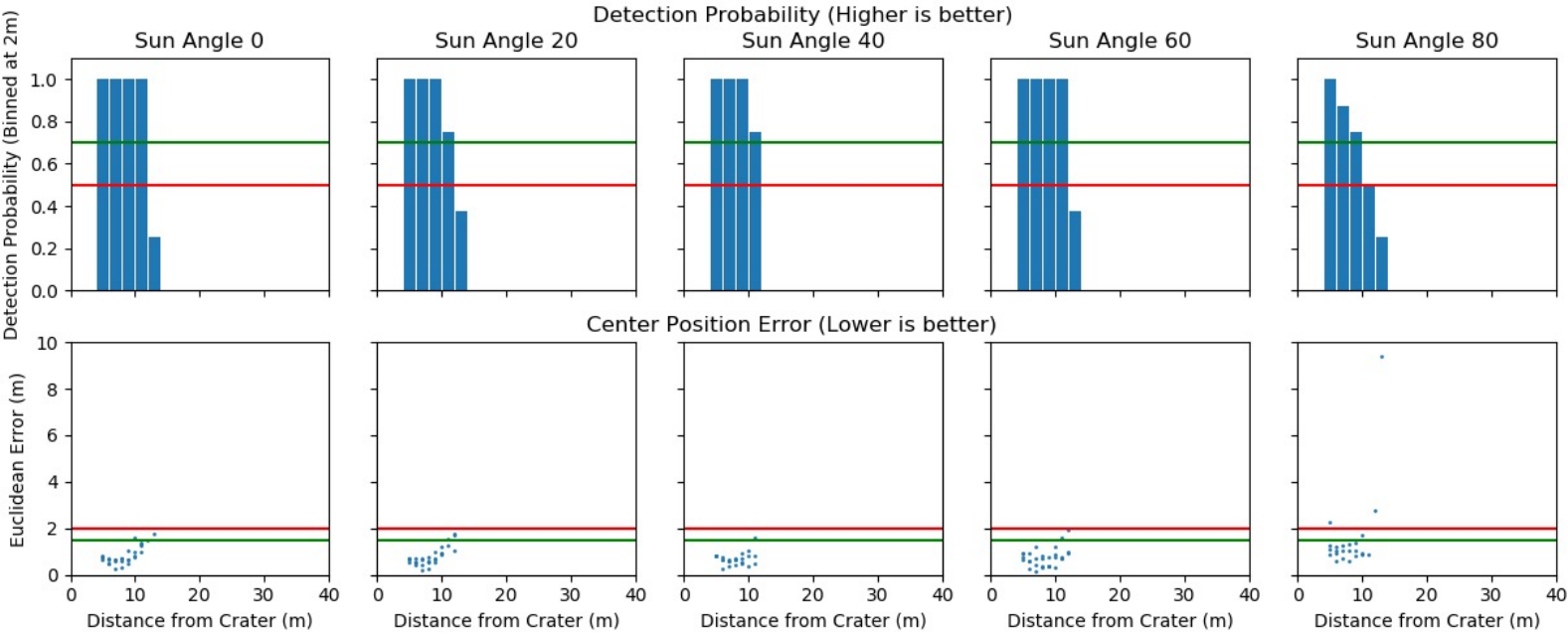}
    \caption{\textbf{Stereo-based detection results for a 5 m diameter crater (simulated images).}}
    \label{fig:stereo-eval1}
\end{figure*}

\begin{figure*}[!t]
    \centering
    \includegraphics[width=\linewidth]{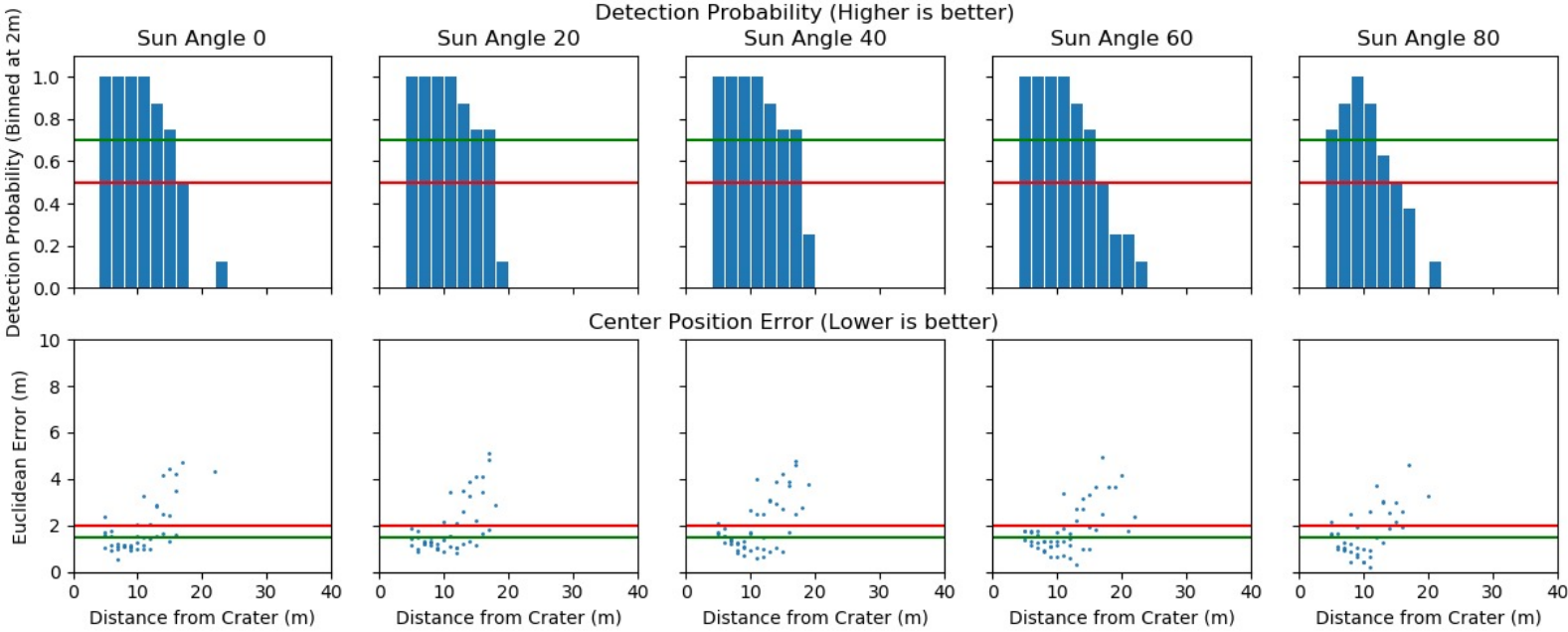}
    \caption{\textbf{Stereo-based detection results for a 10 m diameter crater (simulated images).}}
    \label{fig:stereo-eval2}
\end{figure*}

\begin{figure*}[!t]
    \centering
    \includegraphics[width=\linewidth]{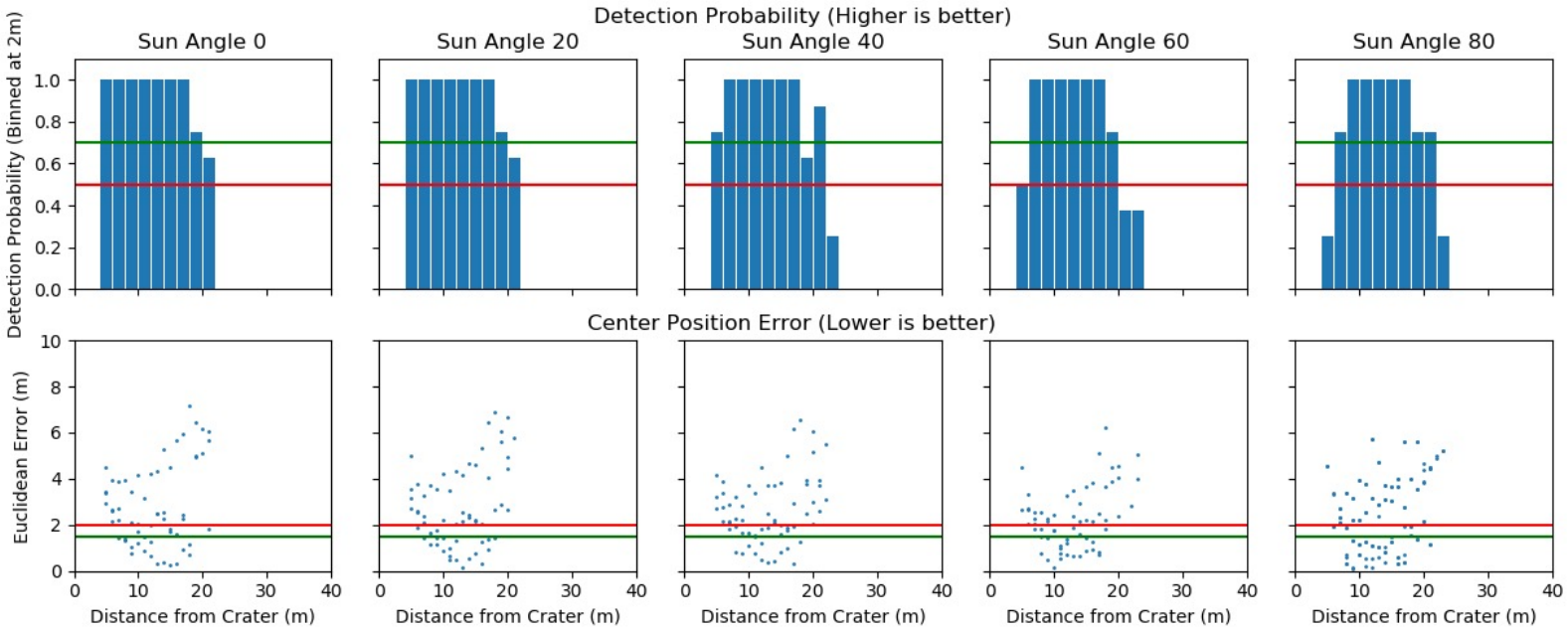}
    \caption{\textbf{Stereo-based detection results for a 15 m diameter crater (simulated images).}}
    \label{fig:stereo-eval3}
\end{figure*}

\begin{figure*}[!t]
    \centering
    \includegraphics[width=\linewidth]{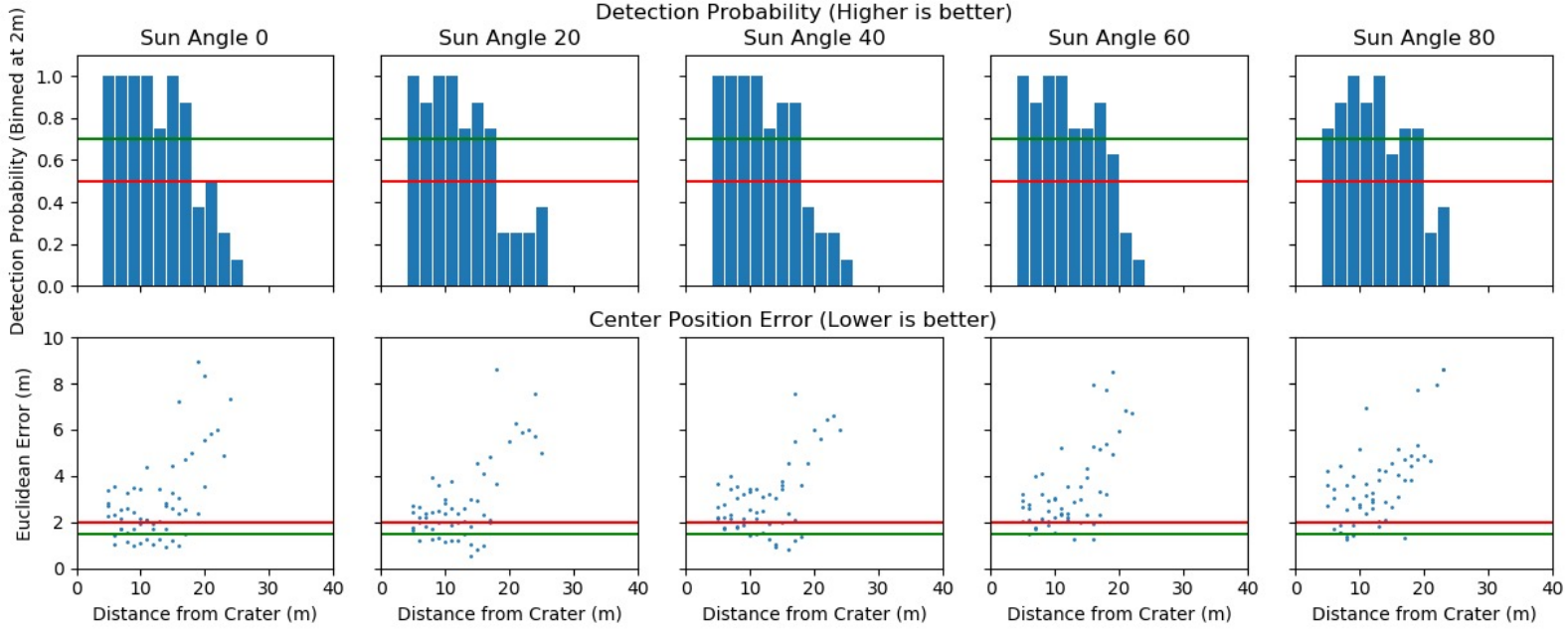}
    \caption{\textbf{Stereo-based detection results for a 20 m diameter crater (simulated images).}}
    \label{fig:stereo-eval4}
\end{figure*}

\begin{table}[!t]
    \centering
    \caption{\textbf{Stereo-based crater position estimation error.}}
    \renewcommand{\arraystretch}{1.5}%
    \begin{tabular}{|c|c|}
    \hline
    \textbf{Crater Diameter (m)} & \textbf{$3\sigma$ Position Error (m)} \\
    \hline
    5 & 1.02 \\
    10 & 3.22 \\
    15 & 4.48 \\
    20 & 4.35 \\
    \hline
    \end{tabular}
    \label{tab:stereo-table}
\end{table}

\subsection{Performance Evaluation for Stereo Approach}
An example of a qualitative stereo-based crater detection result is shown in Figure \ref{fig:stereo-qual1}. In the disparity map, white pixels represent missing data; this often happens at range discontinuities, which occur in this image mainly at the near and/or far rims of craters. Figures \ref{fig:stereo-eval1}, \ref{fig:stereo-eval2}, \ref{fig:stereo-eval3}, \ref{fig:stereo-eval4} show the results of the stereo crater detection algorithm with craters of diameter 5 to 20 m, sun zenith angles from 0$^{\circ}$ to 80$^{\circ}$, ranges from the crater near rim from 5 to 20 m, and 4 different approach angles (not plotted separately). As with the LIDAR detection algorithm, we see the highest accuracy in crater position estimation at smaller crater sizes, with performance becoming less precise as the size of the crater increases. Detection performance drops off at shorter ranges with the stereo-based algorithm than with LIDAR, probably due to greater noise in the range data. $P_d$ reaches the KPP goal level for craters $\geq$ 10 m in diameter (i.e. to ranges $\geq$ 15 m), and for 5 m diameter craters to a range of 12 m.

Furthermore, the stereo detection algorithm appears to be robust to different lighting conditions – there are no significant differences between detection probability nor detected position accuracy across the different sun angles tested. Table 5 shows $3\sigma$ errors in estimated crater positions, computed over all ranges with significant detection rates. These are considerably larger than observed for the LIDAR-based approach, though the KPP goal is met (3$\sigma$  error < 1.5 m for 5 m craters at range up to 15 m). From the position error graphs, error grows rapidly with increasing range to the crater, which is consistent with the range error characteristics of stereo vision. The error observed here should be sufficient for rover navigation; nevertheless, future work will address reducing the error.

\begin{figure*}[!htb]
    \centering
    \begin{tabular}{cc}
    \includegraphics[width=0.48\linewidth]{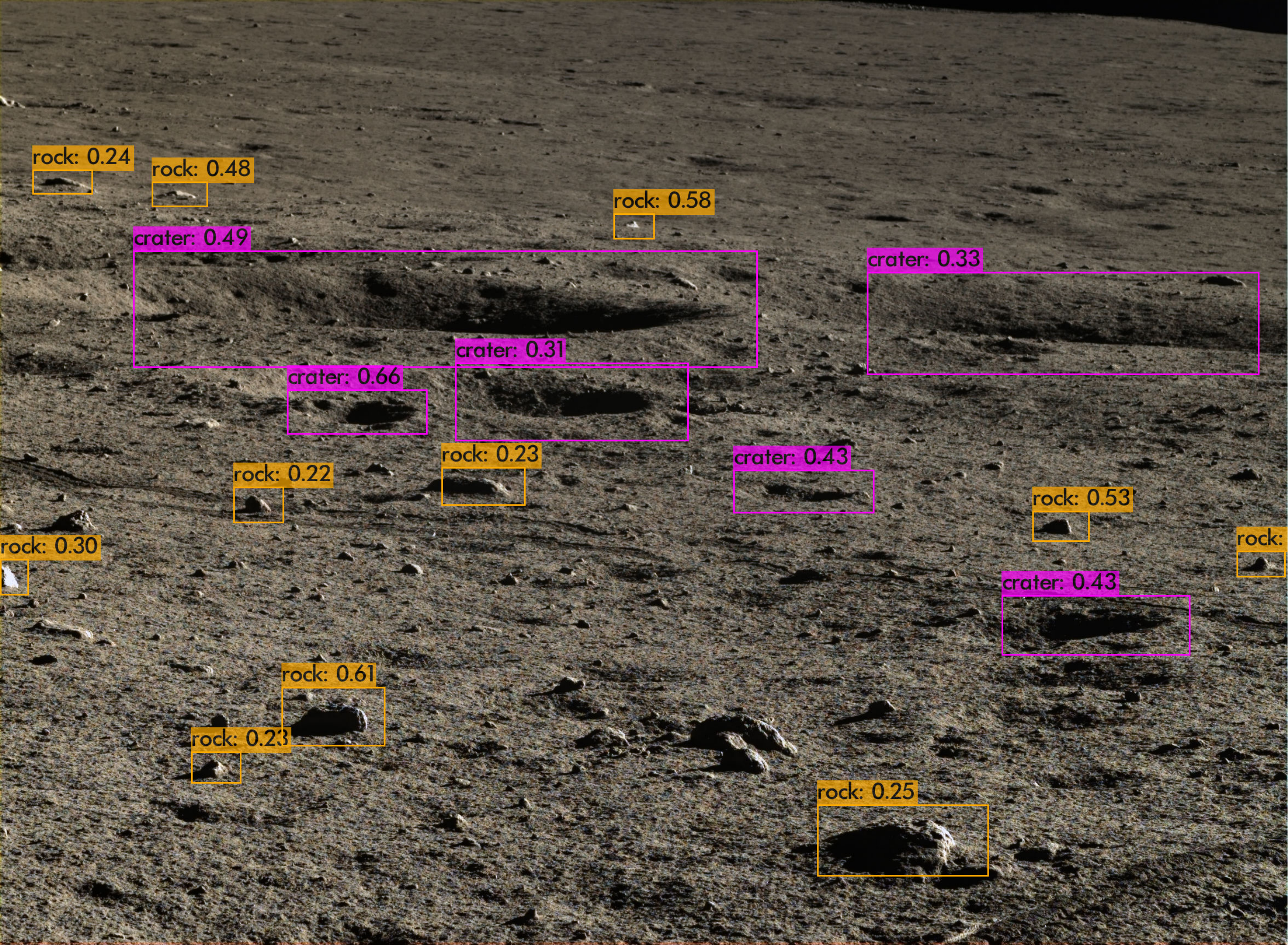} &
    \includegraphics[width=0.48\linewidth]{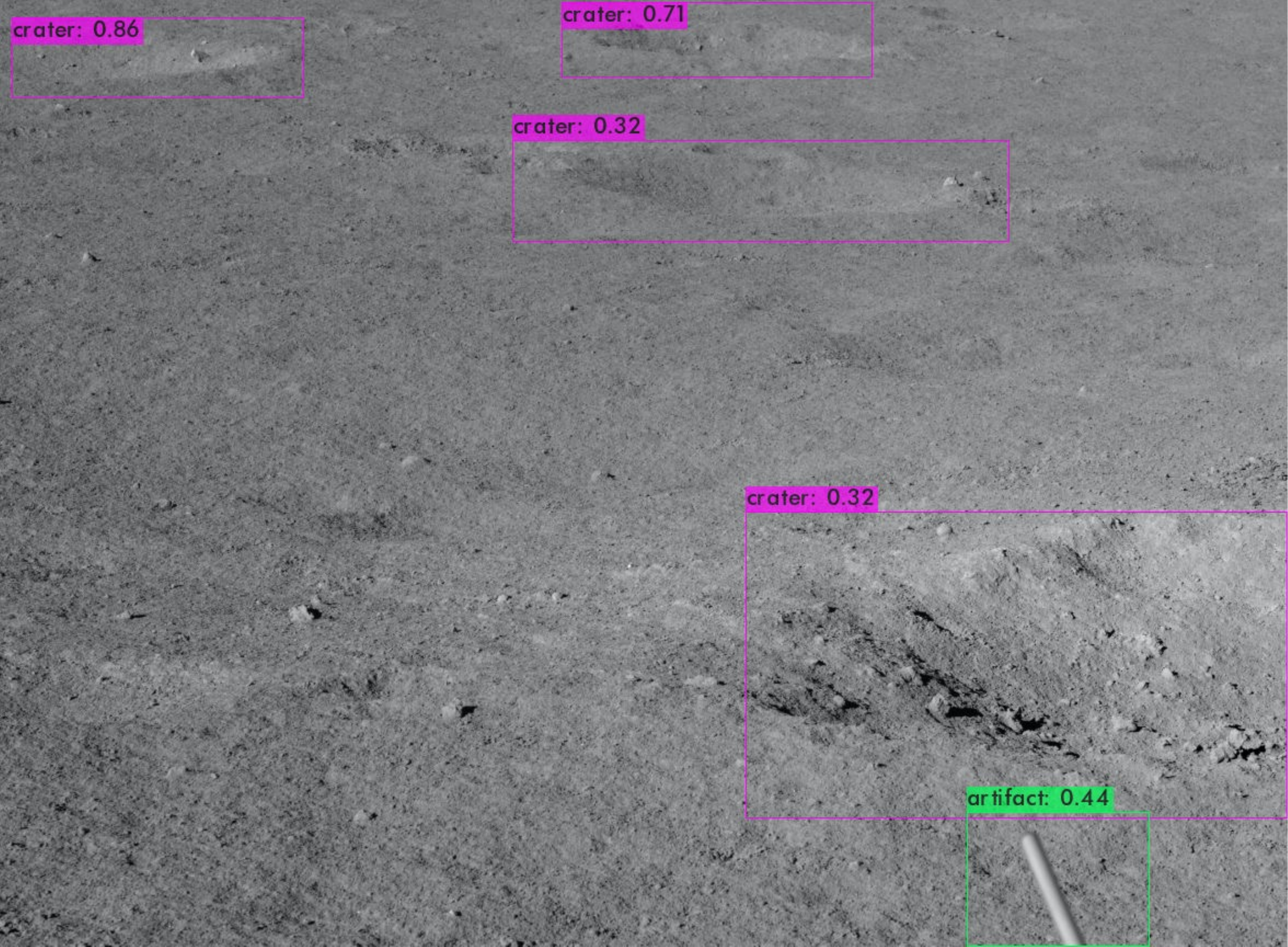} \\
    \end{tabular}
    \caption{\textbf{Qualitative Results demonstrating crater detection on lunar images from Chang’e 3 and Chang’e 4 mission.}}
    \label{fig:mono-qual}
\end{figure*}

\subsection{Performance Evaluation for Monocular Approach}
We evaluated the machine-learned model’s ability to do “transfer learning” to other datasets by qualitatively evaluating the performance on a large dataset of real lunar images from the Chang'e rovers. Example qualitative results in Figure \ref{fig:mono-qual} show that the model was able to successfully detect craters over a range of parameters (illumination conditions, distance to crater, and crater size). As with the other algorithms, quantitative performance evaluation was done with the simulated dataset. At this time, the monocular approach only outputs crater detections in image space; it does not yet use 3D data to estimate crater positions or diameters in 3D. Therefore, performance evaluation assessed the probability of crater detection ($P_d$) in image space as a function of variable crater size and crater distance.

\begin{figure*}[!htb]
    \centering
    \begin{tabular}{c c}
    \includegraphics[width=0.4\linewidth]{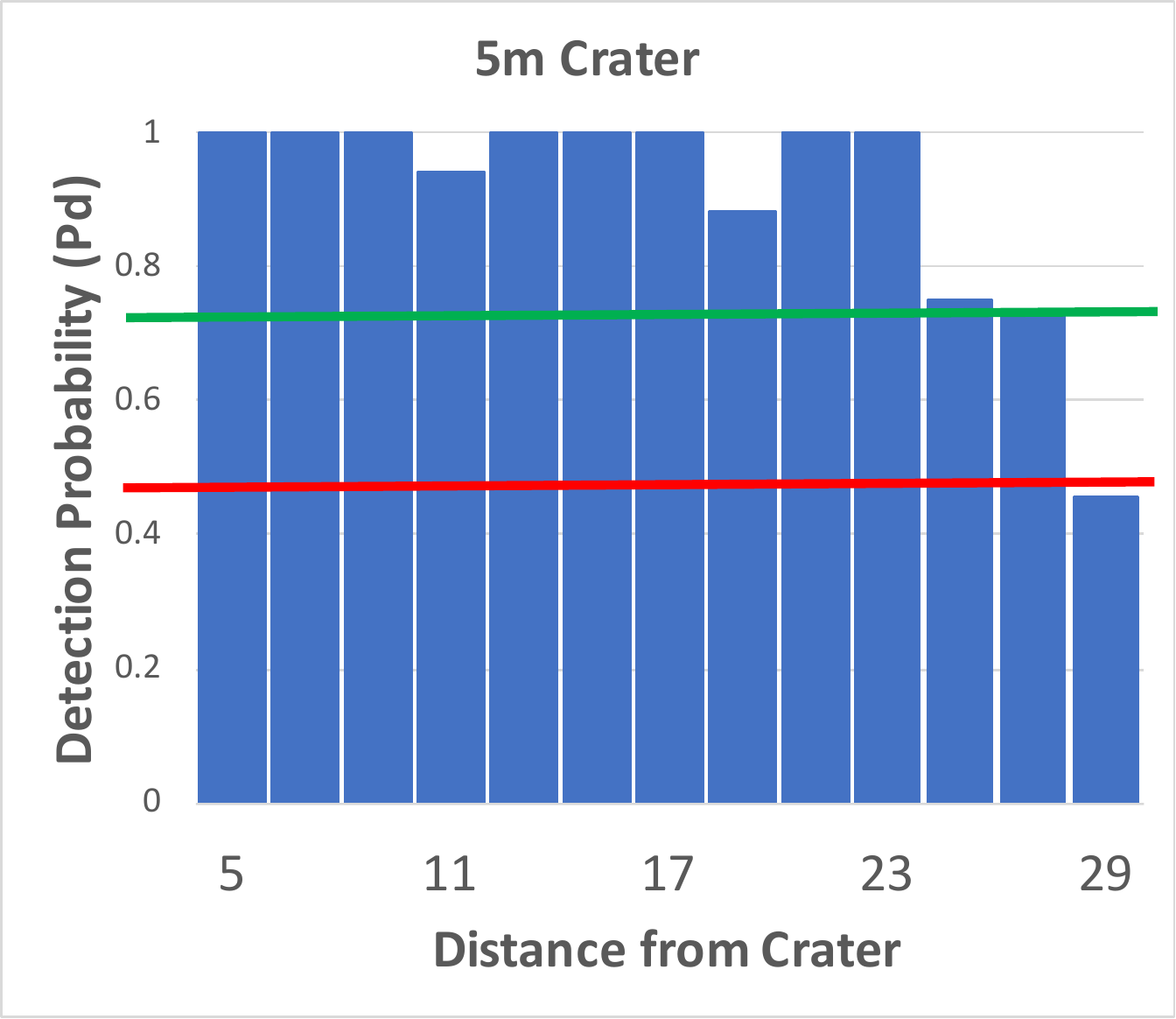} &
    \includegraphics[width=0.4\linewidth]{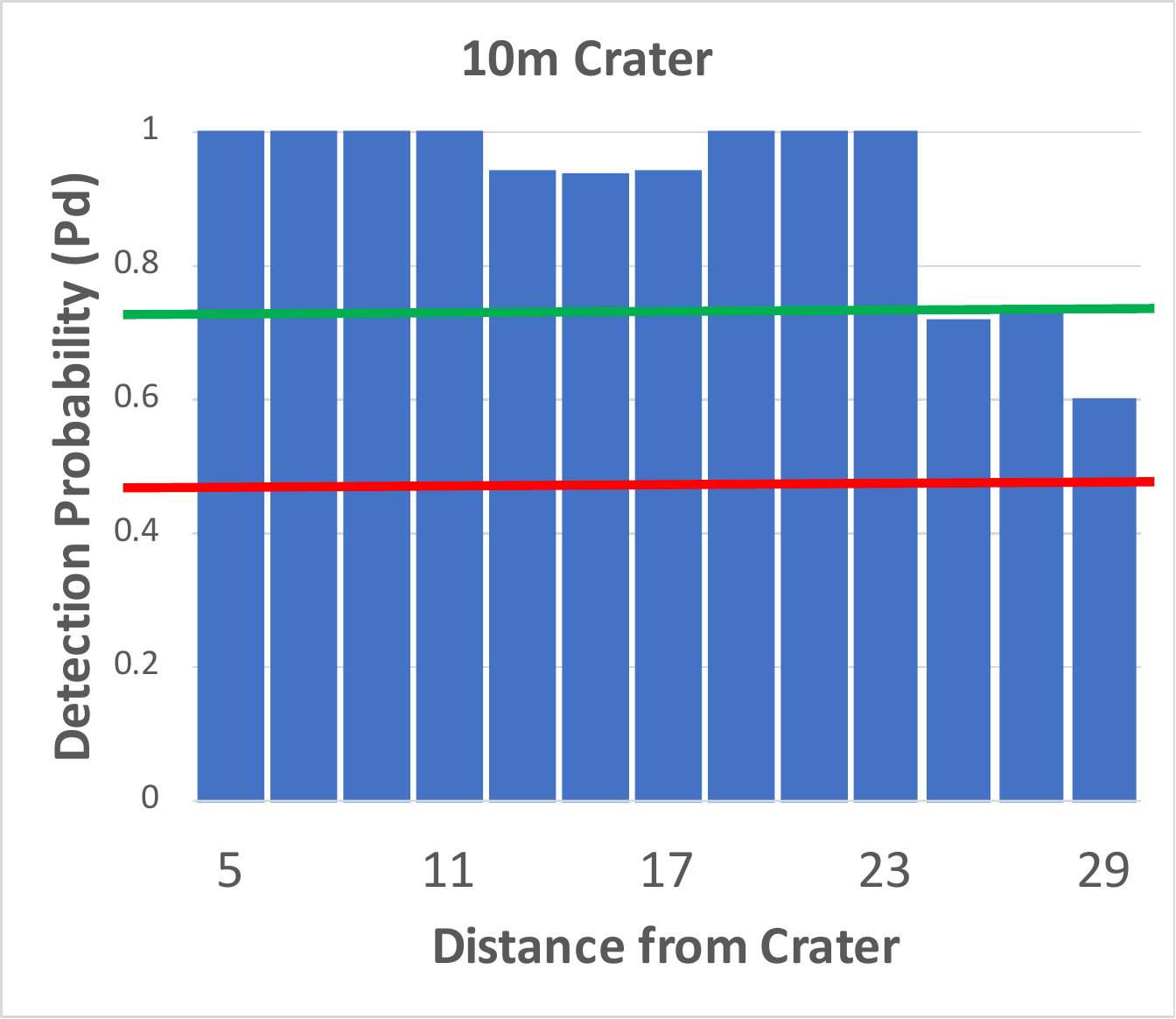} \\
    \includegraphics[width=0.4\linewidth]{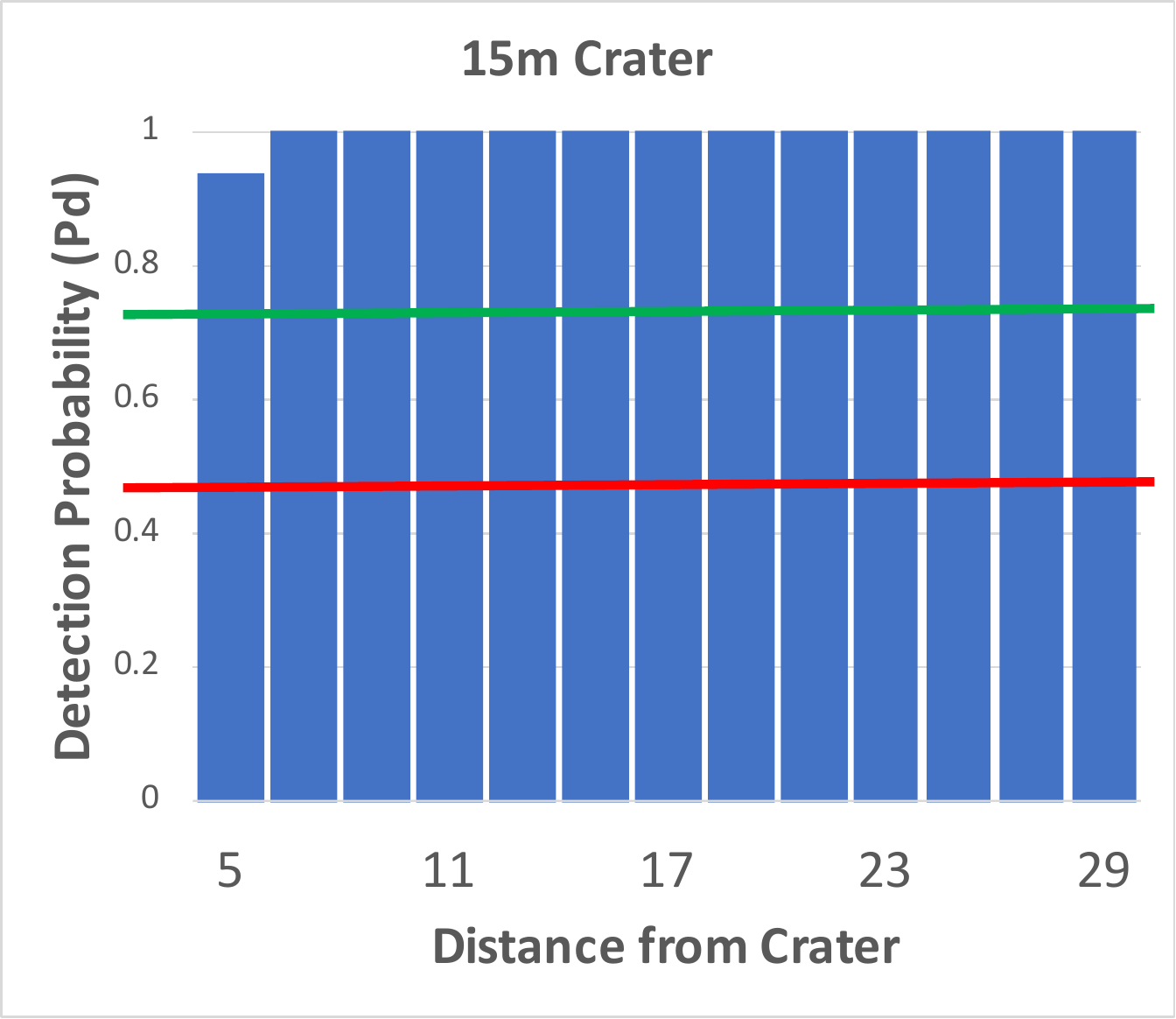} &
    \includegraphics[width=0.4\linewidth]{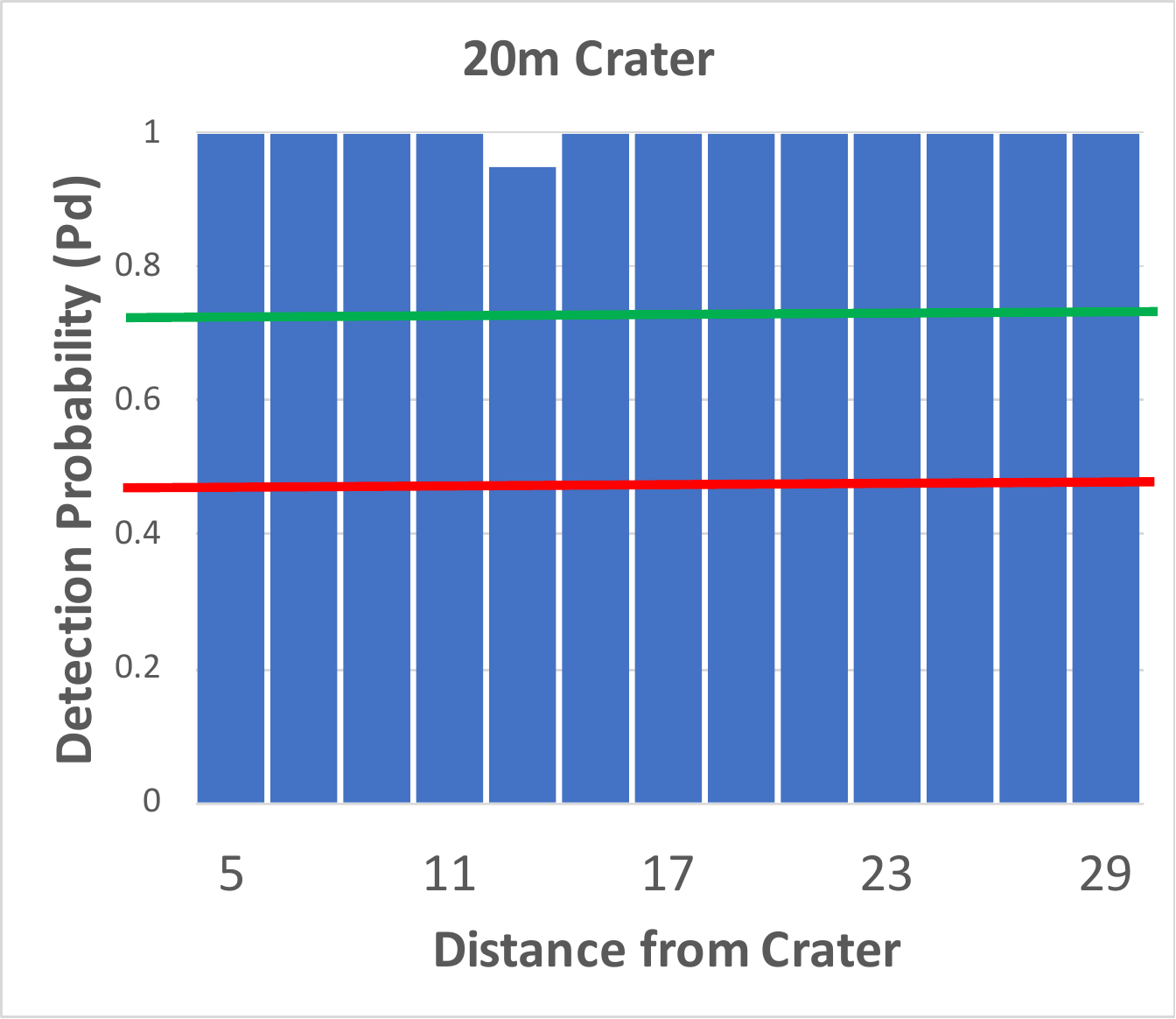} \\
    \end{tabular}
    \caption{\textbf{Monocular crater detection performance (simulated images).}}
    \label{fig:mono-eval}
\end{figure*}

Figure \ref{fig:mono-eval} plots the results for 4 different craters sizes ranging from 5 to 20 m, at 2 m crater distance intervals ranging from 5 m to 29 m. For craters larger than 10 m diameter, Pd is very high for the entire range of crater distances. For craters less than 10 m in diameter, $P_d$ is very high $P_d$ for crater distance up to 25 m, but drops beyond that range. This is expected, because it gets progressively more difficult to reliably detect small craters at larger distances. In all cases, detection is good at greater ranges than have been achieved to date with geometric analysis of 3D point cloud data. This could be due to the nature of image data (i.e. better spatial resolution), to the power of CNN-based detection algorithms compared to manually-designed geometric detection algorithms, or both.

\begin{figure*}[!htb]
    \centering
    \begin{tabular}{c c}
    \includegraphics[width=0.45\linewidth]{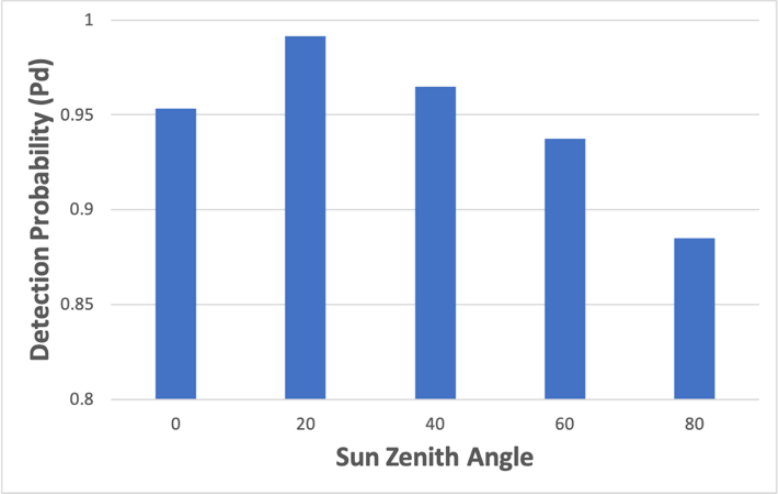} &
    \includegraphics[width=0.45\linewidth]{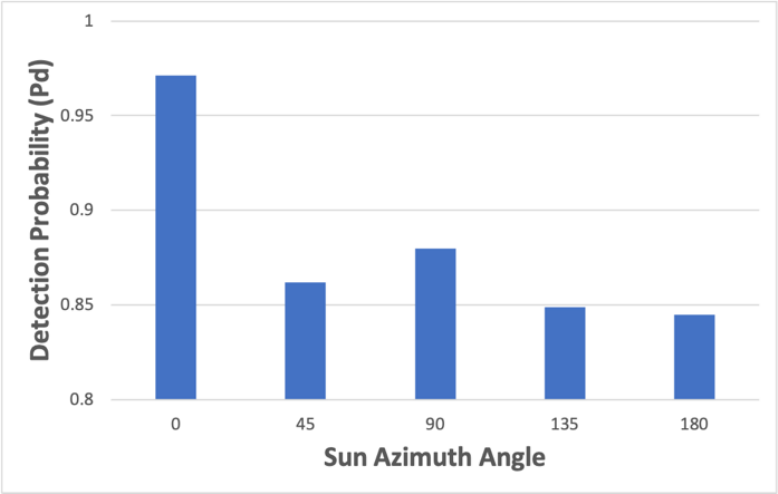} \\
    \end{tabular}
    \caption{\textbf{Monocular detection vs sun zenith and azimuth angles.}}
    \label{fig:mono-eval2}
\end{figure*}

We also studied the performance of crater detection as a function of illumination condition. Figure \ref{fig:mono-eval2} shows the performance of crater detection (averaged over all crater sizes and distances) as a function of sun zenith and azimuth angles. Sun zenith angles range from 0$^{\circ}$ (at zenith) to 80$^{\circ}$ (near the horizon). We observe that crater detection performance decreases with increasing sun zenith angle; we believe this is due to the harsh illumination conditions (e.g. long shadows) when the sun is close to the horizon. It is interesting to note that the performance at 0$^{\circ}$ is worse than at 20$^{\circ}$. We believe this can be attributed to the fact that when the sun is directly overhead, there are no shadows at all, which can sometimes serve as discriminative features at crater rims. Overall, the detection probability for all sun zenith angles is still above the goal of 75\%. Sun azimuth angles were tested from 0$^{\circ}$ to 180$^{\circ}$, where 0$^{\circ}$ corresponds to the sun being directly in front of the camera and 180$^{\circ}$ has the sun is exactly behind the camera. Crater detection performance is highest when the sun is directly in front of the rover and worst when the sun is directly behind the rover. This is the sun in front of the camera casts visible shadows, whereas the sun behind the camera creates the opposition effect \cite{hapke1993opposition}, which reduces image contrast at low phase angles.

\begin{figure}[!htb]
    \centering
    \includegraphics[width=\linewidth]{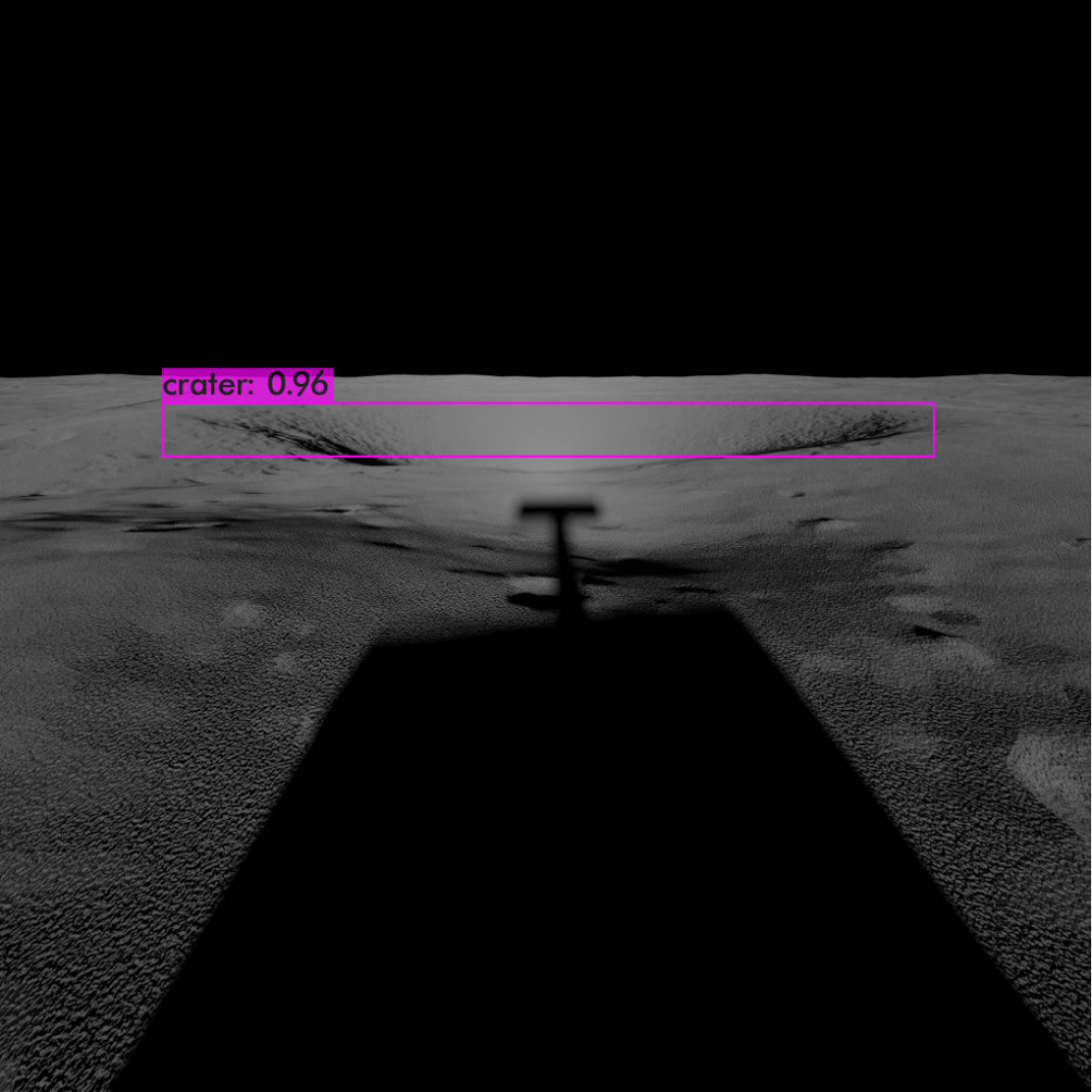}
    \caption{\textbf{Monocular detection example in a case with the opposition effect.}}
    \label{fig:mono-qual2}
\end{figure}

Figure \ref{fig:mono-qual2} shows a qualitative example of successful crater detection even in the presence of opposition effect. For the given example, we observe a 20 m crater at 10 m distance from the rover; sun zenith angle is 80$^{\circ}$ and azimuth is 180$^{\circ}$. As can be seen from the example, the opposition effect significantly reduces the contrast of the crater rim sections closest and farthest from the camera; in this example, there were still shadows on the sides of the crater.

\section{Discussion}
\label{sec:discussion}
In this paper, we have presented initial results of a planned three-year project to develop techniques for lunar rover absolute localization using craters as landmarks. This year, we developed algorithms for onboard crater detection using three methods based on (1) 3D point clouds obtained from lidar, (2) 3D point clouds obtained from  stereo vision, and (3) pattern recognition with monocular images. Development and testing was done with real lunar images from the rovers in the Chang’e 3 and 4 missions, and quantitative evaluation of statistical performance metrics was done with simulated data, generated using a Blender-based lunar scene simulator.

Results of quantitative performance evaluation with geometric analysis applied to simulated 3D point clouds from LIDAR show high reliability for detecting craters with a leading edge within about 15 m from the rover. The results also suggest that rover localization with an error less than 5 m is highly probable. Somewhat simpler geometric analysis methods were applied to simulated 3D point clouds from stereo vision, which are noisier than LIDAR-based point clouds. Stereo-based detection degraded at shorter range than for LIDAR and obtained significantly higher crater position estimation error; nevertheless, rover localization with error in the 5 m range still appears to be possible. Monocular appearance-based detection was done with a CNN-based machine learning algorithm; to date, this produces detection results in image space, but does not produce 3D crater position and size estimates. Detection performance exceeded the other two methods, making this a very promising approach.

Future work will acquire a large-scale dataset of real images from an analog crater site at Cinder Lake crater field near Flagstaff, AZ, improve the detection algorithms and experiment with combinations of the approaches, and address rover position estimation by matching crater detections with craters in the landmark database.


\acknowledgments
The research was carried out at the Jet Propulsion Laboratory, California Institute of Technology, under a contract with the National Aeronautics and Space Administration (80NM0018D0004).

\bibliographystyle{IEEEtran}
\bibliography{references}

\thebiography
\begin{biographywithpic}
{Larry Matthies}{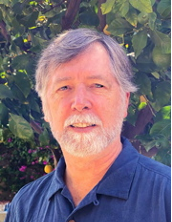}
received B.S., M. Math, and PhD degrees in Computer Science from the University of Regina (1979), University of Waterloo (1981), and Carnegie Mellon University (1989). He has been with JPL for more than 32 years. He has conducted technology development in perception systems for autonomous navigation of robotic vehicles for land, sea, air, and space. He supervised the JPL Computer Vision group for 21 years. He led development of computer vision algorithms for Mars rovers, landers, and helicopters. He is a Fellow of the IEEE and a member of the editorial boards of Autonomous Robots and the Journal of Field Robotics.
\end{biographywithpic}

\begin{biographywithpic}
{Shreyansh Daftry}{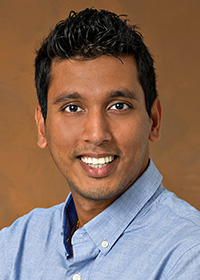}
is a Robotics Technologist at NASA Jet Propulsion Laboratory, California Institute of Technology. He received his M.S. degree in Robotics from the Robotics Institute, Carnegie Mellon University in 2016, and his B.S. degree in Electronics and Communication Engineering in 2013. His research interest lies is at intersection of space technology and autonomous robotic systems, with an emphasis on machine learning applications to perception, planning and decision making. At JPL, he has worked on mission formulation for Mars Sample Return, and technology development for autonomous navigation of ground, airborne and subterranean robots.
\end{biographywithpic}

\begin{biographywithpic}
{Scott Tepsuporn}{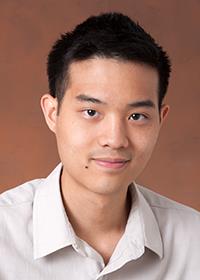}
Scott Tepsuporn is a Robotics Electrical Engineer within the Mobility and Robotic Systems section at the Jet Propulsion Laboratory where he is involved in the design and implementation of various motion planning, computer vision, and mission autonomy work. He received his B.S. degrees from the University of Virginia in Computer Science and Electrical Engineering.
\end{biographywithpic}

\begin{biographywithpic}
{Yang Cheng}{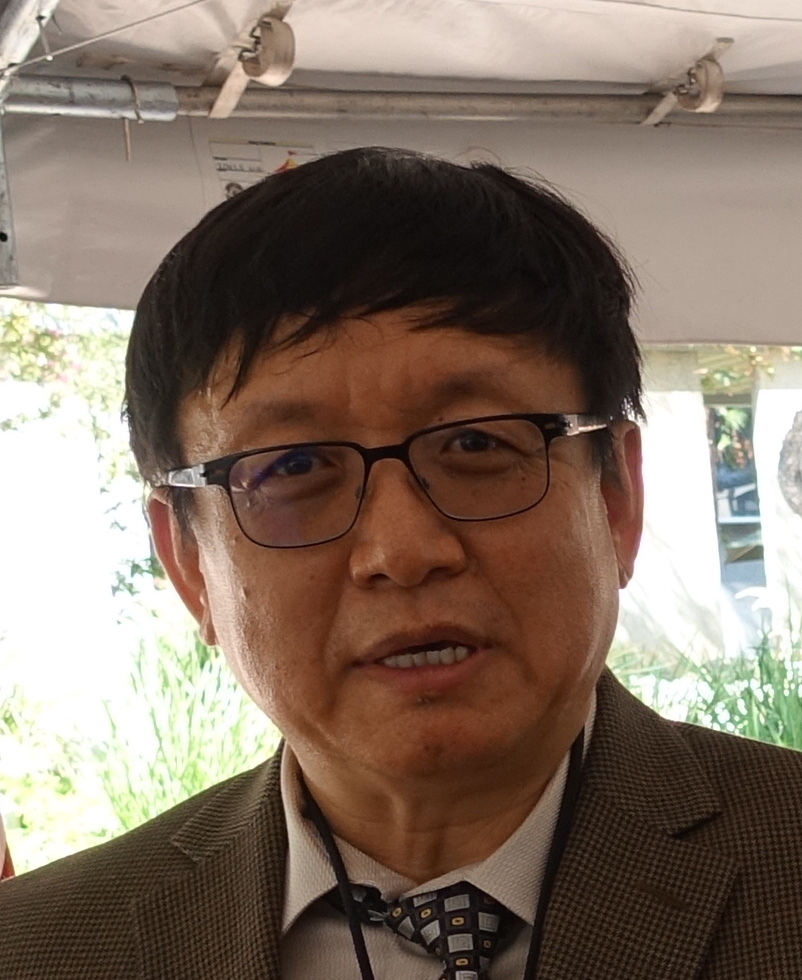}
is a principal member of the Aerial System Perception Group at JPL.  Dr. Cheng is a leading expert in the area of optical spacecraft navigation, terrain relative navigation, surface robotic perception and cartography. He has made significant technical contributions in technology advancement in the area of structure from motion, 3D surface reconstruction, surface hazard detection and mapping for spacecraft landing site selection, stereo matching and sub-pixel interpolation, map projection .   Dr. Cheng is the key  developer for MER descent image motion estimation system (DIMES) and Mars2020 lander vision system (LVS) and  the first ever onboard reference map for Mars2020 LVS.
\end{biographywithpic}

\begin{biographywithpic}
{Deegan Atha}{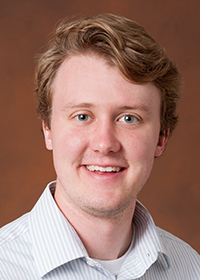}
is a Robotics Technologist with the Mobility and Robotic Systems section at the Jet Propulsion Laboratory. He received his B.S. degree from Purdue University in Electrical Engineering. His interests include machine learning, computer vision, and multi-agent systems. 
\end{biographywithpic}

\begin{biographywithpic}
{R. Michael Swan}{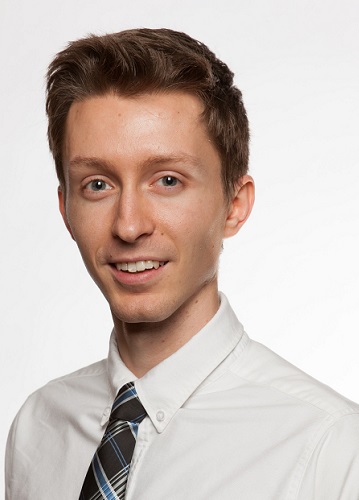}
is a Robotics Systems Engineer at NASA Jet Propulsion Laboratory,
California Institute of Technology. He received his B.S. in Computer Engineering from Walla Walla University and his M.S. in Computer Science from the University of Southern California. He is interested in robotic surface and aerial autonomy, perception, simulation, and robotic system architecture.
\end{biographywithpic}

\begin{biographywithpic}
{Sanjna Ravichandar}{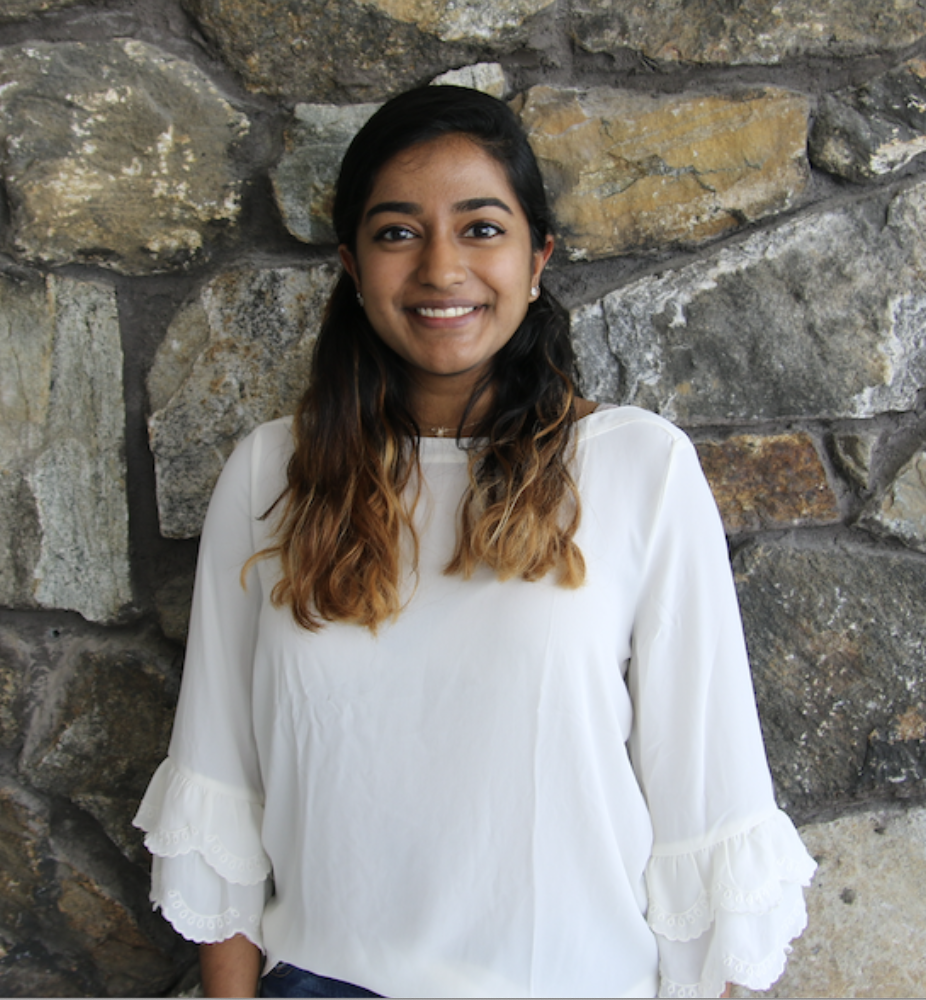}
is an undergraduate student at Massachusetts Institute of Technology (MIT) pursuing Electrical Engineering and Computer Science in the class of 2022. Her research focuses in machine learning and robotics, and she works at MIT's Computer Science and Artificial Intelligence Laboratory in learning methods for dexterous manipulation and reinforcement learning controllers for legged robots. Select past projects include developing image classifiers for JPL's PDS Image Atlas, implementing gait normalization algorithms for active prosthetics, and working on automated root cause analysis for IBM Research, and developing flight control applications for NASA Johnson Space Center.
\end{biographywithpic}

\begin{biographywithpic}
{Masahiro Ono}{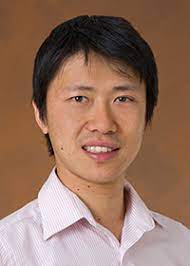}
is the Group Leader of the Robotic Surface Mobility Group. Since he joined JPL in 2013, he has led a number of research projects on Mars rover autonomy, as well as three NIAC studies on Enceladus Vent Explorer and Comet Hitchhiker. Hiro was a flight software developer of M2020’s Enhanced AutoNav and the lead of M2020 Landing Site Traversability Analysis. He also led the development of a machine learning-based Martian terrain classifier, SPOC (Soil Property and Object Classification), which won JPLs Software of the Year Award in 2020. 
\end{biographywithpic}

\end{document}